%% file: emnlp2021.tex
\pdfoutput=1

\documentclass[11pt]{article}

\usepackage{emnlp2021}

\usepackage{times}
\usepackage{latexsym}

\usepackage[T1]{fontenc}

\usepackage[utf8]{inputenc}

\usepackage{microtype}

\usepackage{bm}
\usepackage{bbm}
\usepackage{graphicx}
\usepackage{color}
\usepackage{multicol}
\usepackage{multirow}
\usepackage{wrapfig,lipsum,booktabs}
\usepackage[ruled,vlined,linesnumbered]{algorithm2e}
\usepackage{pgfplots}
\usepackage{filecontents}
\usepackage{xcolor}
\usepackage{tikz}
\usepackage{xspace}
\usepackage{makecell}
\usepackage{soul}
\usepackage{amsmath,amsfonts,amssymb}
\usepackage{subcaption}
\usetikzlibrary{calc}
\usepgfplotslibrary{groupplots}
\usetikzlibrary{angles,quotes} 
\usepackage{hyperref}
\usepackage{cleveref}
\usepackage{paralist}
\usepackage{cancel}
\usepackage{xspace}
\usepackage{todonotes}
\usepackage{tabu}
\usepackage{rotating}
\usepackage{etoolbox}

\newcommand{\dtrain}{D_{\textrm{trn}}}
\newcommand{\dvalid}{D_{\textrm{dev}}}
\newcommand{\dtest}{D_{\textrm{tst}}}
\newcommand{\vx}{\pmb{x}}
\newcommand{\vy}{\pmb{y}}

\newcommand{\vtheta}{\pmb{\theta}}
\newcommand{\vscorer}{\pmb{\psi}}

\newcommand{\avgbleu}{\mu_{\textrm{BLEU}}}
\newcommand{\model}[1]{\textsc{#1}\xspace}
\newcommand{\ours}{\model{MultiUAT}}
\newcommand{\dds}{\model{MultiDDS-S}}

\newcommand*{\affmark}[1][*]{\textsuperscript{#1}}

\DeclareMathOperator*{\argmin}{argmin}
\DeclareMathOperator*{\argmax}{argmax}

\title{Uncertainty-Aware Balancing for \\ Multilingual and Multi-Domain Neural Machine Translation Training}

\author{
  Minghao Wu\affmark[$\heartsuit$]\thanks{{} {} Work done during the internship at Huawei Noah's Ark Lab.}  \qquad Yitong Li\affmark[$\clubsuit$] \qquad Meng Zhang\affmark[$\spadesuit$] \qquad Liangyou Li\affmark[$\spadesuit$] \\  {\bf Gholamreza Haffari\affmark[$\heartsuit$]} \qquad {\bf Qun Liu\affmark[$\spadesuit$]} \\
  \affmark[$\heartsuit$]Monash University \quad \affmark[$\clubsuit$] Huawei Technologies Co., Ltd. \quad \affmark[$\spadesuit$] Huawei Noah's Ark Lab \\
  \texttt{\{minghao.wu, gholamreza.haffari\}@monash.edu} \\
  \texttt{\{liyitong3, zhangmeng92, liliangyou, qun.liu\}@huawei.com}
}

\begin{document}
\renewcommand{\tableautorefname}{Table}
\renewcommand{\sectionautorefname}{Section}
\renewcommand{\subsectionautorefname}{Section}
\renewcommand{\figureautorefname}{Figure}
\renewcommand{\equationautorefname}{Equation}
\renewcommand{\algorithmautorefname}{Algorithm}
\newcommand{\linenoautorefname}{Line}

\maketitle
\begin{abstract}
Learning multilingual and multi-domain translation model is challenging as the heterogeneous and imbalanced data make the model converge inconsistently over different corpora in real world.
One common practice is to adjust the share of each corpus in the training, so that the learning process is balanced and low-resource cases can benefit from the high-resource ones.
However, automatic balancing methods usually depend on the intra- and inter-dataset characteristics, which is usually agnostic or requires human priors.
In this work, we propose an approach, \textbf{\ours}, that dynamically adjusts the training data usage based on the model's uncertainty on a small set of trusted clean data for multi-corpus machine translation.
We experiment with two classes of uncertainty measures on multilingual (16 languages with 4 settings) and multi-domain settings (4 for in-domain and 2 for out-of-domain on English-German translation) and demonstrate our approach \ours substantially outperforms its baselines, including both static and dynamic strategies.
We analyze the cross-domain transfer and show the deficiency of static and similarity based methods.\footnote{Code available at \url{https://github.com/huawei-noah/noah-research/tree/master/noahnmt/multiuat}}
\end{abstract}

\input{1_introduction.tex}
\input{2_preliminaries.tex}
\input{3_method.tex}
\input{4_setup.tex}
\input{5_results.tex}

\input{6_analysis.tex}
\input{7_related_work.tex}
\input{8_conclusion.tex}

\bibliography{anthology,custom}
\bibliographystyle{acl_natbib}

\clearpage
\appendix
\input{appendix_1_results.tex}
\input{appendix_2_hyperparams.tex}

\end{document}

%% file: 1_introduction.tex
\section{Introduction}
\label{sec:intro}
Text corpora are commonly collected from several different sources in different languages, raising the problem of learning a NLP system from the heterogeneous corpora, such as multilingual models \citep{wu-dredze-2019-beto, DBLP:journals/corr/abs-1907-05019, aharoni-etal-2019-massively, freitag-firat-2020-complete, arthur-etal-2021-multilingual} and multi-domain models \citep{daume-iii-2007-frustratingly, li-etal-2019-semi, DBLP:conf/emnlp/DengYYDL20, jiang-etal-2020-multi}. 
A strong demand is to deploy a unified model for all the languages and domains, because a unified model is much more resource-efficient, and knowledge learned from high-resource languages/domains (HRLs/HRDs) can be transferred to low-resource languages/domains (LRLs/LRDs).

One common issue on training models across corpora is that data from a variety of corpora are both heterogeneous (different corpora reveal different linguistic properties) and imbalance (the accessibility of training data varies across corpora). 
The standard practice to address this issue is to adjust the training data distribution heuristically by up-sampling the training data from LRLs/LRDs \citep{DBLP:journals/corr/abs-1907-05019, conneau-etal-2020-unsupervised}. 

\citet{DBLP:journals/corr/abs-1907-05019} rescale the training data distribution with a heuristic temperature term and demonstrate that the ideal temperature can substantially improve the overall performance. 
However, the optimal value for such heuristics is both hard to find and varies from one experimental setting to another \citep{wang-neubig-2019-target,DBLP:conf/icml/WangPMACN20,wang-etal-2020-balancing}. 
\citet{DBLP:conf/icml/WangPMACN20} and \citet{wang-etal-2020-balancing} hypothesize that the training data instances that are similar to the validation set can be more beneficial to the evaluation performance and propose a general reinforcement-learning framework \textit{Differentiable Data Selection} (DDS) that automatically adjusts the importance of data points, whose reward is the cosine similarity of the gradients between a small set of trusted clean data and training data.
They instantiate this framework on multilingual NMT, known as \textsc{MultiDDS}, to dynamically weigh the importance of language pairs.
Both the hypothesis and the proposed approach rely on the assumption that knowledge learned from one corpus can always be beneficial to the other corpora.

However, their assumption does not always hold. 
If the knowledge learned from one corpus is not able to be transferred easily or is useless to the other corpora, this approach fails.
Unlike cosine similarity, \textit{model uncertainty} is free from the aforementioned assumption on cross-corpus transfer.
From a Bayesian viewpoint, the model parameters can be considered as a random variable that describes the dataset. 
If one dataset is well-described by the model parameters, its corresponding model uncertainty is low, and vice versa. 
This nature makes the model uncertainty an ideal option to weigh the datasets.

In this work, we propose an approach \ours that leverages the model uncertainty as the reward to dynamically adjust the sampling probability distribution over multiple corpora. 
We consider the model parameter as a random variable that describes the multiple training corpora. 
If one corpus is well-described by the model compared with other corpora, we spare more training efforts to the other poorly described corpora. 
We conduct extensive experiments on multilingual NMT (16 languages with 4 settings) and multi-domain NMT (4 for in-domain and 2 for out-of-domain), comparing our approach with heuristic static strategy and dynamic strategy. 
In multilingual NMT, we improve the overall performance from +0.83 BLEU score to +1.52 BLEU score among 4 settings, comparing with the best baseline. 
In multi-domain NMT, our approach improves the  in-domain overall performance by +0.58 BLEU score comparing with the best baseline and achieves the second best out-of-domain overall performance. 
We also empirically illustrate the vulnerability of cosine similarity as the reward in the training among multiple corpora.

%% file: 2_preliminaries.tex
\section{Preliminaries}
\label{sec:prelim}
\paragraph{Standard NMT} A standard NMT model, parameterized by $\vtheta$, is commonly trained on one language pair $\dtrain^o=\{(\vx, \vy)_i\}^{M}_{i=1}$ from one domain.
The objective is to minimize the negative log-likelihood of the training data with respect to $\vtheta$:
\begin{equation}
        {\mathcal{L}_{s}(\dtrain^o; \vtheta)} = -\sum_{i=1}^{M} \log p(\vy | \vx; \vtheta) \, .
    \label{eq:singleloss}
\end{equation}

\paragraph{Multi-corpus NMT} Both multilingual NMT and multi-domain NMT can be summarized as multi-corpus NMT that aims to build a unified translation system to maximize the overall performance across all the language pairs or domains.
Formally, let us assume we are given a number of datasets $\dtrain=\{\dtrain^j\}^{N}_{j=1}$ from $N$ languages pairs or domains, in which $\dtrain^j = \{ (\vx, \vy)^{j}_{i} \}^{M_j}_{i=1}$, where ${M_j}$ is size of $j$-th language/domain.
Similar to \autoref{eq:singleloss}, a simple way of training multi-corpus NMT model is to treat all instances equally:
\begin{equation}
    \mathcal{L}(\dtrain; \vtheta) = \sum^{N}_{j=1} \mathcal{L}_{s}(\dtrain^j; \vtheta) \, .
    \label{eq:multiloss}
\end{equation}

\paragraph{Heuristic strategy for multi-corpus training}
In practice, \autoref{eq:multiloss} can be reviewed as training using mini-batch sampling according to the proportion of these corpora, $q(n) = \frac{M_n}{\sum_{i=1}^{N} M_i}$, and thus we minimize:
\begin{equation}
    \mathcal{L}(\dtrain; \vtheta, q(n)) = \mathbb{E}_{n \sim q(n)} \left[\mathcal{L}_s(\dtrain^n ; \vtheta)\right] \, .
\end{equation}
However, this simple training method does not work well in real cases, where low-resource tasks are under-trained.

A heuristic static strategy is to adjust the proportion exponentiated by a temperature term $\tau$ \citep{DBLP:journals/corr/abs-1907-05019, conneau-etal-2020-unsupervised}:
\begin{equation}
    q_\tau(n) = \frac{q(n)^{1/\tau}}{\sum_{i=1}^{N} q(i)^{1/\tau}} \, .
    \label{eq:temp}
\end{equation}
And the loss function for multi-corpus training can be re-formulated as:
\begin{equation}
    \mathcal{L}(\dtrain; \vtheta, q_\tau(n)) = \mathbb{E}_{n \sim q_\tau(n)} [\mathcal{L}_s(\dtrain^n; \vtheta)] \, .
    \label{eq:multiloss-exp}
\end{equation}
Specifically, $\tau = 1$ or $\tau = \infty$ is equivalent to proportional (\autoref{eq:multiloss}) or uniform sampling respectively.

\paragraph{Differentiable Data Selection (DDS)}
\citet{DBLP:conf/icml/WangPMACN20} propose a general framework that automatically weighs training instances to improve the performance while relying on an independent set of held-out data $\dvalid$.
Their framework consists of two major components, the model $\vtheta$ and the scorer network $\vscorer$.
The scorer network $\vscorer$ is trained to assign a sampling probability to each training instance, denoted as $p_{\vscorer}(\vx, \vy)$, based on its contribution to the validation performance.
The training instance that contributes more to the validation performance is assigned a higher probability and more likely to be used for updating the model $\vtheta$.
This strategy aims to maximize the overall performance over $\dvalid$ and is expected to generalize well on the unseen $\dtest$ with the assumption of the independence and identical distribution between $\dvalid$ and $\dtest$.
Therefore, the objective can formulated as:
\begin{equation}
    \begin{aligned}
        \vscorer &= \argmin_{\vscorer} \mathcal{L}(\dvalid; \vtheta(\vscorer))\\
        \vtheta(\vscorer) &= \argmin_{\vtheta}\mathbb{E}_{(\vx, \vy) \sim p_{\vscorer}(\vx, \vy)} [\mathcal{L}(\dtrain ; \vtheta)] \, ,
    \end{aligned}
    \label{eq:ddsobj}
\end{equation}
and $\vscorer$ and $\vtheta$ are updated iteratively using \textit{bilevel optimization} \citep{DBLP:journals/anor/ColsonMS07, von2011market}.

%% file: 3_method.tex
\section{Methodology}

In this work, we leverage the idea of DDS under the multi-corpus scenarios.
We utilize a differentiable domain/language scorer $\vscorer$ to weigh the training corpora.
To learn $\vscorer$, we exploit the model uncertainty to measure the model's ability over the target corpus.
Below, we elaborate on the details of our method.

\subsection{Model Uncertainty}
\textit{Model uncertainty} can be a measure that indicates whether the model parameters $\vtheta$ are able to describe the data distribution well \citep{DBLP:conf/nips/KendallG17, dong-etal-2018-confidence, DBLP:conf/aaai/XiaoW19}.
Bayesian neural networks can be used for quantifying the model uncertainty \citep{DBLP:journals/compsys/BuntineW91}, which models the $\vtheta$ as a probabilistic distribution with constant input and output.

From the Bayesian point of view, $\vtheta$ is interpreted as a random variable with the prior $p(\vtheta)$.
Given a dataset $D$, the posterior $p(\vtheta|D)$ can be obtained via Bayes' rule.
However, the exact Bayesian inference is intractable for neural networks, so that it is common to place the approximation $q(\vtheta)$ to the true posterior $p(\vtheta|D)$.
Several variational inference methods have been proposed \citep{DBLP:conf/nips/Graves11, pmlr-v37-blundell15, DBLP:conf/icml/GalG16}.

In this work, we leverage Monte Carlo Dropout \citep{DBLP:conf/icml/GalG16} to obtain samples of sentence-level translation probability.
To quantify the model uncertainty when the model makes predictions, we treat the sentence-level translation probability as random variable.
We run $K$ forward passes with a random subset of model parameters $\vtheta$ deactivated, which is equivalent to drawing samples from the random variable, and average the samples as the estimate of the model uncertainty.\footnote{$K$ is set to 30 in our work.}
Consider an ensemble of models $\{p_{\vtheta_k}(\vy|\vx)\}^{K}_{k=1}$ sampled from the approximate posterior $q(\vtheta)$, the predictive posterior can be obtained by taking the expectation over multiple inferences:
\begin{equation}
    \begin{aligned}
    \label{equ:mcd}
        p(\vy|\vx, D) &\approx \mathbb{E}_{\vtheta \sim q(\vtheta)} [p(\vy|\vx, \vtheta)] \\
                      &\approx \frac{1}{K}\sum_{k=1}^{K} p_{\vtheta_{k}}(\vy|\vx) \, .
    \end{aligned}
\end{equation}

\subsection{Uncertainty-Aware Training}

\begin{algorithm}[t]
    \SetKwInOut{Input}{Input}
    \SetKwInOut{Output}{Output}
    \Input{$D^n = \{ (\vx_{m}^{n}, \vy_{m}^{n}) \}^{M_n}_{m=1}$, $N$ corpora with the size of $M_n$ for the $n$-th corpus; $S$, update frequency of $\vscorer$; $J$, total training steps;}
    \Output{The converged model $\vtheta$}
    \SetAlgoLined
    Initialize $p_{\vscorer}(n;\dtrain)$ as \autoref{eq:temp} with $\tau=1$\; \label{line:line1}
    \For{i=0 to J}{
        $\tilde{n} \sim p_{\vscorer}(n)$\;
        sample batch $(\vx, \vy) \sim \dtrain^{\tilde{n}}$\;
        $\vtheta \leftarrow \vtheta - \alpha \cdot \nabla_{\vtheta} \mathcal{L}(\vy|\vx ; \vtheta) $\;
        \If{i \% S == 0}{
            \For{n=1 to N}{
                $(\vx', \vy') \sim \dvalid^n$\;
                Compute reward/uncertainty $R(n)$ for $\dvalid^n$ as in \autoref{sec:uncertainmeasure}\;
            }
            $\vscorer \leftarrow \vscorer - \sum_{n=1}^{N}R(n) \cdot \nabla_{\vscorer} \log p_{\vscorer}(n)$\;
        }
    }
    \caption{Training with \ours}
    \label{alg:multiuat}
\end{algorithm}

To make the training more efficient and stable, \ours leverages the scorer network $\vscorer$ to dynamically adjust the sampling probability distribution of domains/languages.\footnote{We parameterize the scorer $\vscorer$ following \citet{wang-etal-2020-balancing}.}

We present the pseudo-code for training with \ours in \autoref{alg:multiuat}.
\ours firstly parameterizes the initial sampling probability distribution for multi-corpus training with $\vscorer$ as \autoref{eq:temp} with the warm-up temperature $\tau=1$.
For the computational efficiency, the scorer network $\vscorer$ is updated for every $S$ steps.
When updating $\vscorer$, we randomly draw one mini-batch from each validation set $\{\dvalid^i\}_{i=1}^{N}$ and compute the corresponding uncertainty measure as in \autoref{sec:uncertainmeasure} with Monte Carlo Dropout to approximate the model uncertainty towards this corpus, assuming the validation set is representative enough for its corresponding true distribution. 
The corpus associated with high uncertainty is considered to be relatively poorly described by the model $\vtheta$ and its sampling probability will be increased.
The model $\vtheta$ is updated by mini-batch gradient descent between two updates of $\vscorer$, like common gradient-based optimization, and hence the objective is formulated as follows:
\begin{equation}
    \begin{aligned}
        \vscorer &= \argmin_{\vscorer} {\mathcal{L}(\dvalid; \vtheta(\vscorer))}  \\
        \vtheta(\vscorer) &= \argmin_{\vtheta}\mathbb{E}_{n \sim p_{\vscorer}(n)} [\mathcal{L}(\dtrain^n;\vtheta)] \, .
    \end{aligned}
\end{equation}

A considerable problem here is \autoref{eq:ddsobj} is not directly differentiable w.r.t. the scorer $\vscorer$.
To tackle this problem, reinforcement learning (RL) with suitable reward functions is required \cite{DBLP:conf/emnlp/FangLC17,DBLP:conf/icml/WangPMACN20}:
\begin{equation}
    \vscorer \leftarrow \vscorer - \sum_{n=1}^{N}R(n) \cdot \nabla_{\vscorer} \log p_{\vscorer}(n) \, .
\end{equation}
Details for the reward functions $R(n)$ are depicted at \autoref{sec:uncertainmeasure} and the update of $\vscorer$ follows the REINFORCE algorithm \citep{10.1007/BF00992696}.

\subsection{Uncertainty Measures}
\label{sec:uncertainmeasure}
We explore the utility of two groups of model uncertainty measures: \textit{probability-based} and \textit{entropy-based} measures at the sentence level \citep{wang-etal-2019-improving-back,fomicheva-etal-2020-unsupervised,malinin2021uncertainty}.

\paragraph{Probability-Based Measures} We explore four probability-based uncertainty measures  following the definition of \citet{wang-etal-2019-improving-back}.
For the sampled model parameters $\vtheta_k$, with the teacher-forced decoding, we note the predicted probability of the $t$-th position as:
\begin{equation}
    \hat{y}_{n,t} = \argmax_y p(y|\vx_{n}, \vy_{n,<t};\vtheta_k) \, ,
\end{equation}
where we have used the ground truth prefix $\vy_{n,<t}$ in the conditioning context.
We then define the reward function as the following uncertainty measures:
\begin{itemize}
    \item
\textit{Predicted Translation Probability} (\textsc{PreTP}): The predicted probability of the sentence,
    \begin{equation} \nonumber
        R_{\text{\textsc{PreTP}}}(n;\vtheta_k) = 1 - \prod_{t=1}^{T} p(\hat{y}_{n,t}|\vx_{n}, \vy_{n,<t};\vtheta_k) \, .
    \end{equation}
    \item
\textit{Expected Translation Probability} (\textsc{ExpTP}): The expectation of the distribution of maximal position-wise translation probability,
    \begin{equation} \nonumber
        R_{\text{\textsc{ExpTP}}}(n;\vtheta_k) = 1 - \mathbb{E} \left[ p(\hat{y}_{n,t} |\vx_{n}, \vy_{n,<t};\vtheta_k) \right] \, .
    \end{equation}
    \item
\textit{Variance of Translation Probability} (\textsc{VarTP}): The variance of  the distribution of maximal position-wise translation probability,
    \begin{equation} \nonumber
        R_{\text{\textsc{VarTP}}}(n;\vtheta_k) = \text{Var}[p(\hat{y}_{n,t}|\vx_{n}, \vy_{n,<t};\vtheta_k)] \, .
    \end{equation}
    \item
\textit{Combination of Expectation and Variance} (\textsc{ComEV}):
    \begin{equation} \nonumber
        R_{\text{\textsc{ComEV}}}(n;\vtheta_k) = \frac{\text{Var}[p(\hat{y}_{n,t}|\vx_{n}, \vy_{n,<t};\vtheta_k)]}{\mathbb{E} [p(\hat{y}_{n,t}|\vx_{n}, \vy_{n,<t};\vtheta_k)]} \, .
    \end{equation}
\end{itemize}

\paragraph{Entropy-Based Measures} \citet{malinin2021uncertainty} consider the uncertainty estimation for auto-regressive models at the token-level and sequence-level and treat the entropy of the posterior as the total uncertainty in the prediction of $\vy$. 
Following their interpretation, we leverage the \textit{entropy} as the measure of the model uncertainty.

Drawing a pair of sentence $(\vx, \vy)$ with $T$ target tokens from the $n$-th corpus $D^n$, the reward function is defined as the averaged entropy over all the positions:
\begin{equation}
    R(n;\vtheta) = \frac{1}{T} \sum_{t=1}^{T} \sum_{v=1}^{\mathcal{V}} p(y_{n,t,v})\log p(y_{n,t,v}) \, .
    \label{eq:uncertainty}
\end{equation}
where $\mathcal{V}$ is the vocabulary size and $p(y_{n,t,v})$ stands for the predicted conditional probability $p(y_{n,t,v}|\vx, \vy_{n,<t,v};\vtheta_k)$ on the $v$-th word in the vocabulary.

In this work, we explore the utility of two entropy-based uncertainty measures as follows:
\begin{itemize}
    \item \textit{Entropy of the sentence} (\textsc{EntSent}): The average entropy of the sentence  as defined in \autoref{eq:uncertainty}.
    \item \textit{Entropy of EOS} (\textsc{EntEOS}): The entropy of the symbol EOS in the sentence as defined in \autoref{eq:uncertainty} where $t=T$.
\end{itemize}

Following \autoref{equ:mcd}, we have the final reward by multiple sampled $\vtheta_k$ for each uncertainty reward respectively: 
\begin{equation}
    R(n) = \frac{1}{K} \sum_{k=1}^{K} R_{\{.\}}(n;\vtheta_k) \, .
\end{equation}

%% file: 4_setup.tex
\section{Experimental Setup}

\subsection{Baselines}
We compare \ours with both static and dynamic strategies as follows:

\paragraph{Heuristics} We run experiments with proportional (\textsc{Prop.}, $\tau=1$), temperature (\textsc{Temp.}, $\tau = 5$) and uniform (\textsc{Uni.}, $\tau = \infty$) in \autoref{eq:temp} following \citet{wang-etal-2020-balancing}.

\paragraph{\dds} We compare with the best model \dds proposed by \citet{wang-etal-2020-balancing} over multilingual NMT tasks.
Its reward for the $n$-th corpus is defined using cosine similarity:
\begin{equation}
    \begin{aligned}
       & R_{\cos}(n)
      =\frac{1}{N} \sum_{i=1}^{N} \cos(\nabla_{\vtheta} \mathcal{L}(\dvalid^i), \nabla_{\vtheta} \mathcal{L}(\dtrain^n)) \, .
    \end{aligned}
    \label{eq:cosinereward}
\end{equation}

\subsection{Multilingual Setup}
We follow the identical setup as \citet{wang-etal-2020-balancing} in the multilingual NMT.
The model is trained on two sets of language pairs based on the language diversity.

\paragraph{Related} 4 LRLs (Azerbaijani: \texttt{aze}, Belarusian: \texttt{bel}, Glacian: \texttt{glg}, Slovak: \texttt{slk}) and a related HRL for each LRL (Turkish: \texttt{tur}, Russian: \texttt{rus}, Portuguese: \texttt{por}, Czech: \texttt{ces}).

\paragraph{Diverse} 8 languages with varying amounts of data, picked without consideration for relatedness (Bosnian: \texttt{bos}, Marathi: \texttt{mar}, Hindi: \texttt{hin}, Macedonian:\texttt{mkd}, Greek: \texttt{ell}, Bulgarian: \texttt{bul}, French: \texttt{fra}, Korean: \texttt{kor}).

We run many-to-one (M2O, translating 8 languages to English) and one-to-many (O2M, translating English to 8 languages) translations for both diverse and related setups.\footnote{Refer to \citet{wang-etal-2020-balancing} for dataset statistics.}

\subsection{Multi-Domain Setup}

We run experiments on English-German translation and collect six corpora from WMT2014 \citep{bojar-etal-2014-findings} and the Open Parallel Corpus \citep{tiedemann-2012-parallel}, 4 for in-domain and 2 for out-of-domain:

\paragraph{In-Domain (ID)}  (i) \texttt{WMT}, from WMT2014 translation task \citep{bojar-etal-2014-findings} with the concatenation from \texttt{newstest2010} to \texttt{newstest2013} for validation and \texttt{newstest2014} for testing; (ii) \texttt{Tanzil},\footnote{\url{https://opus.nlpl.eu/Tanzil.php}} a collection of Quran translations; (iii) \texttt{EMEA},\footnote{\url{https://opus.nlpl.eu/EMEA.php}} a parallel corpus from the European Medicines Agency; (iv) \texttt{KDE},\footnote{\url{https://opus.nlpl.eu/KDE4.php}} a parallel corpus of KDE4 localization files.

\paragraph{Out-Of-Domain (OOD)} (i) \texttt{QED},\footnote{\url{https://opus.nlpl.eu/QED.php}} a collection of subtitles for educational videos and lectures \citep{abdelali-etal-2014-amara}; (ii) \texttt{TED},\footnote{\url{https://opus.nlpl.eu/TED2013.php}} a parallel corpus of TED talk subtitles. These two domains are only used for out-of-domain evaluation.

All these corpora are first tokenized by Moses \citep{koehn-etal-2007-moses} and processed into sub-word units by BPE \cite{sennrich-etal-2016-neural} with $32K$ merge operations. 
Sentence pairs that are duplicated and violates source-target ratio of $1.5$ are removed.
The validation sets and test sets are randomly sampled, except for \texttt{WMT}.
The dataset statistics are listed in \autoref{tab:multidomainstat}.

\begin{table}[!t]
\small
    \centering
    \begin{tabular}{llccc}
                          & & Train & Valid & Test \\
    \toprule
    \multirow{4}{*}{ID}
    &\texttt{WMT}                    & $3,950K$ & $11K$   & $3K$   \\ 
    &\texttt{Tanzil}                 & \phantom{00}$449K$  & \phantom{0}$3K$    & $3K$   \\ 
    &\texttt{EMEA}                   & \phantom{00}$277K$  & \phantom{0}$3K$    & $3K$   \\ 
    &\texttt{KDE}                    & \phantom{00}$135K$  & \phantom{0}$3K$    & $3K$   \\
    \midrule
    \multirow{2}{*}{OOD}
    &\texttt{QED}                    & -     & -    & $3K$   \\ 
    &\texttt{TED}                    & -     & -    & $3K$   \\ \bottomrule
    \end{tabular}
    \caption{Dataset statistics of multi-domain corpora.}
    \label{tab:multidomainstat}
\end{table}

\subsection{Model Architecture}
We believe all the approaches involved in this work, including the baseline approaches and \ours, are model-agnostic.
To validate this idea, we experiment two variants of transformer \citep{DBLP:conf/nips/VaswaniSPUJGKP17}.
For multilingual NMT, the model architecture is a transformer with $4$ attention heads and $6$ layers.\footnote{Signature:\\\texttt{multilingual\_transformer\_iwslt\_de\_en}}
And for multi-domain NMT models, we use the standard transformer-base with 8 attention heads and 6 layers.\footnote{Signature: \texttt{transformer}}
All the models in this work are implemented by fairseq \citep{ott-etal-2019-fairseq}.

\subsection{Evaluation}
\label{sec:eval}
We report detokenized BLEU \cite{papineni-etal-2002-bleu} using SacreBLEU \cite{post-2018-call} with statistical significance given by \citet{koehn-2004-statistical}.\footnote{Signature: \texttt{BLEU+case.mixed+numrefs.1\\+smooth.exp+tok.13a+version.1.4.14}} $\avgbleu$ is the macro average of BLEU scores within the same setting, with the assumption that all the language pairs/domains are equally important.

%% file: 5_results.tex
\begin{table*}[!t]
    \small
    \setlength{\tabcolsep}{10pt}
    \centering
    \begin{tabu}{lccccccc}
    
              &     &\multicolumn{4}{c}{Multilingual}       & \multicolumn{2}{c}{Multi-Domain} \\ \cmidrule(lr){3-6} \cmidrule(l){7-8}
                 &     &\multicolumn{2}{c}{Related}     & \multicolumn{2}{c}{Diverse}   & \multirow{2}{*}{ID} & \multirow{2}{*}{OOD} \\
                  \cmidrule(lr){3-4} \cmidrule(lr){5-6}
                 & Average & M2O            & O2M            & M2O            & O2M     &  &       \\
            \toprule
    \textsc{Prop.}\phantom{0}($\tau=1$)      & 22.58 & 24.88          & 15.49          & 26.68          & 16.79   & 35.69 & \textbf{25.15}        \\ 
    \textsc{Temp.} ($\tau=5$)    & 22.80 & 24.00          & 16.61          & 26.01          & 17.94   & 36.92 & 23.46       \\ 
    \textsc{Uni.}\phantom{00}($\tau=\infty$) & 21.86 & 22.63          & 15.54          & 24.81          & 16.86   & 36.85 & 22.22       \\ 
    \dds           & 23.58 & 25.52          & 17.32          & 27.00          & 18.24   & 36.42 & 22.74   \\ \midrule
    \ours & & & & & & & \\
    \phantom{000}+ \textsc{PreTP}       & 24.57\rlap{$\dagger$} & 26.30\rlap{$\dagger$} & 18.36\rlap{$\dagger$} & \underline{27.82}\rlap{$\dagger$} & 19.57\rlap{$\dagger$} & \textbf{37.50}\rlap{$\dagger$}  & 23.77   \\ 
    \phantom{000}+ \textsc{ExpTP}       & \underline{24.62}\rlap{$\dagger$} & \underline{26.36}\rlap{$\dagger$} & \textbf{18.64}\rlap{$\dagger$} & 27.74\rlap{$\dagger$} & 19.59\rlap{$\dagger$} & 37.29  & \underline{23.89}     \\ 
    \phantom{000}+ \textsc{VarTP}       & 24.44\rlap{$\dagger$} & \textbf{26.39}\rlap{$\dagger$}    & 18.58\rlap{$\dagger$}    & \textbf{27.83}\rlap{$\dagger$}    & 19.62\rlap{$\dagger$}  & 35.49 & 23.75        \\ 
    \phantom{000}+ \textsc{ComEV}       & 24.57\rlap{$\dagger$} & 26.26\rlap{$\dagger$}          & 18.58\rlap{$\dagger$}    & 27.75\rlap{$\dagger$}          & 19.67\rlap{$\dagger$}   & 37.25 & 23.36       \\
    \phantom{000}+ \textsc{EntSent}     & \underline{24.62}\rlap{$\dagger$} & 26.34\rlap{$\dagger$}          & \underline{18.63}\rlap{$\dagger$} & 27.71\rlap{$\dagger$}          & \underline{19.68}\rlap{$\dagger$}    & 37.27 & 23.86      \\ 
    \phantom{000}+ \textsc{EntEOS}      & \textbf{24.65}\rlap{$\dagger$} & 26.28\rlap{$\dagger$}          & 18.56\rlap{$\dagger$}          & \underline{27.82}\rlap{$\dagger$}          & \textbf{19.76}\rlap{$\dagger$} & \underline{37.44}\rlap{$\dagger$} & 23.80 \\ \bottomrule

    \end{tabu}
    \caption{$\avgbleu$ for the settings on multilingual and multi-domain NMT. 
    ``Average'' is the macro average of all the BLEU scores for both multilingual and multi-domain settings at \autoref{appsec:comp}.
    Best results are highlighted in \textbf{bold} and second best results are highlighted in \underline{underline}. 
    $\dagger$ indicates the improvement for the corresponding \ours result against \dds result is statistically significant at $p<0.05$ using paired Student's t-test.
    }
    \label{tab:overall}
\end{table*}

\section{Main Results}
The summarized results for both multilingual and multi-domain NMT are presented in \autoref{tab:overall}.\footnote{We report the multilingual results of the heuristic approach and \dds from \citet{wang-etal-2020-balancing} in \autoref{tab:overall}.}
The complete results with statistical significance can be found in \autoref{appsec:comp}.

\paragraph{Multilingual NMT} Overall, dynamic strategies (\dds and \ours) demonstrate their superiority against heuristic static strategies.
As shown in \autoref{tab:overall}, the optimal $\tau$ of heuristic static strategies varies as the combination of corpora changes.
For example, proportional sampling yields best performance on M2O settings, yet achieves the worst performance on O2M settings among heuristic static strategies.
Dynamic strategies are free from adjusting the data usage by tuning the $\tau$.
\dds marginally outperforms heuristic static strategies.
\ours with various uncertainty measures reaches the best performance in all four settings. Based on the detailed results in \autoref{appsec:comp}, we can observe that \ours appears to be more favorable to HRLs.

\paragraph{Multi-domain NMT} \ours outperforms all its baselines on in-domain evaluation and achieves the second best performance on out-of-domain evaluation. \ours with \textsc{PreTP} achieves the optimal balance on in-domain evaluation and the one with \textsc{ExpTP} achieves the second best performance on out-of-domain evaluation.
However, \dds performs poorly on multi-domain NMT and is even outperformed by some heuristic static strategies.

Based on the detailed results in \autoref{appsec:comp}, we can observe that the higher sampling probability for certain domain is commonly but not always positively correlated to the corresponding in-domain performance. 
Uniformly sampling mini-batches from domains does not result in the best performance on LRDs, because the LRDs with too much up-sampling are not able to fully leverage the knowledge from the HRDs.

%% file: 6_analysis.tex
\section{Analysis}

\citet{wang-etal-2020-balancing} conduct exhaustive analyses on multilingual NMT and most of our observations are consistent with theirs.\footnote{The multilingual results of \dds used for analysis are provided by our own implementation with the hyperparameters provided by \citet{wang-etal-2020-balancing}.} Hence, we focus more on analyzing the results on multi-domain NMT.

\subsection{Comparison of Uncertainty Measures}

We explore the utility of different uncertainty measures and display the summarized results in \autoref{tab:overall}.
Different uncertainty measures deliver different results.
We do not observe one uncertainty measure that consistently outperforms others.
The probability-based uncertainty measures seem to be more sensitive to the intra- and inter-dataset characteristics, and perform well on either multilingual NMT or multi-domain NMT.
\ours with the uncertainty measure of \textsc{VarTP} performs substantially worse than other uncertainty measures in multi-domain NMT. 
In contrast to the probability-based uncertainty measures, the entropy-based uncertainty measures are more robust to the change of datasets and deliver relatively stable improvements.
We also find out that \ours with the uncertainty measures demonstrate better out-of-domain generalization in the multi-domain NMT, compared with its baselines.

Based on the detailed results in \autoref{appsec:comp}, \ours with the entropy-based uncertainty measures demonstrates better robustness against the change of datasets. Therefore, we mainly compare \ours with the uncertainty measure of \textsc{EntEOS} against the baselines in the following analyses, based on the macro-average results on both multilingual and multi-domain NMT.

\begin{figure}[!t]
    \centering
    \begin{tikzpicture}
        \begin{groupplot}[
            cycle list name=color list,
            no markers,
            legend columns=4,
            height=3.5cm,
            width = 0.6\linewidth,
            xlabel={\small Iter. (in thous.)},
            ylabel={\small Samp. Prob.},
            x label style={at={(0.5,0.15)}},
            y label style={at={(0.15,0.5)}},
            legend style={,
                at={(-0.15,-0.35)},
                anchor=north,
                legend columns=-1,
                font=\small
            },
            group style={
                group size=2 by 1,
                xlabels at=edge top,
                x descriptions at=edge bottom,
                y descriptions at=edge left,
                horizontal sep=0pt
            },
            tick label style={font=\tiny}
        ]
        \nextgroupplot[
            title={\small \dds},
        ]
        \addplot table [mark=none, x=idx, y=eng-bos, col sep=comma] {multidds-multilingual-diverse-o2m-probs.csv};
        \addplot table [mark=none, x=idx, y=eng-mar, col sep=comma] {multidds-multilingual-diverse-o2m-probs.csv};
        \addplot table [mark=none, x=idx, y=eng-hin, col sep=comma] {multidds-multilingual-diverse-o2m-probs.csv};
        \addplot table [mark=none, x=idx, y=eng-mkd, col sep=comma] {multidds-multilingual-diverse-o2m-probs.csv};
        \addplot table [mark=none, x=idx, y=eng-ell, col sep=comma] {multidds-multilingual-diverse-o2m-probs.csv};
        \addplot table [mark=none, x=idx, y=eng-bul, col sep=comma] {multidds-multilingual-diverse-o2m-probs.csv};
        \addplot table [mark=none, x=idx, y=eng-fra, col sep=comma] {multidds-multilingual-diverse-o2m-probs.csv};
        \addplot table [mark=none, x=idx, y=eng-kor, col sep=comma] {multidds-multilingual-diverse-o2m-probs.csv};

        \nextgroupplot[
            title={\small \ours},
        ]
        \addplot table [mark=none, x=idx, y=eng-bos, col sep=comma] {multiuat-multilingual-diverse-o2m-probs.csv};
        \addplot table [mark=none, x=idx, y=eng-mar, col sep=comma] {multiuat-multilingual-diverse-o2m-probs.csv};
        \addplot table [mark=none, x=idx, y=eng-hin, col sep=comma] {multiuat-multilingual-diverse-o2m-probs.csv};
        \addplot table [mark=none, x=idx, y=eng-mkd, col sep=comma] {multiuat-multilingual-diverse-o2m-probs.csv};
        \addplot table [mark=none, x=idx, y=eng-ell, col sep=comma] {multiuat-multilingual-diverse-o2m-probs.csv};
        \addplot table [mark=none, x=idx, y=eng-bul, col sep=comma] {multiuat-multilingual-diverse-o2m-probs.csv};
        \addplot table [mark=none, x=idx, y=eng-fra, col sep=comma] {multiuat-multilingual-diverse-o2m-probs.csv};
        \addplot table [mark=none, x=idx, y=eng-kor, col sep=comma] {multiuat-multilingual-diverse-o2m-probs.csv};
        \legend{eng-bos,eng-mar,eng-hin,eng-mkd,eng-ell,eng-bul,eng-fra,eng-kor}
        \end{groupplot}
        
    \end{tikzpicture}
    \caption{Iteration (Iter.)-sampling probability of \dds and \ours under the multilingual O2M-diverse setting.}
    \label{fig:multilingualcurve}

\end{figure}
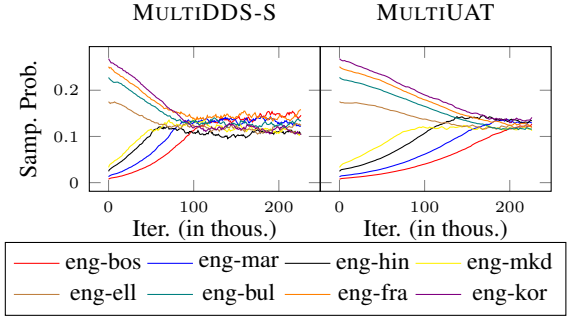

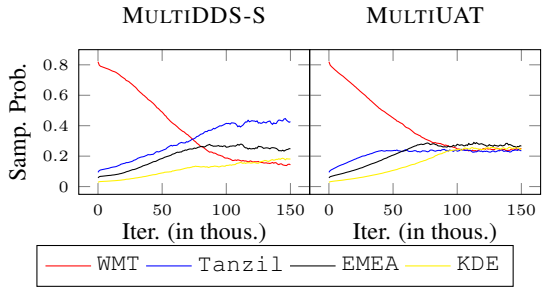
\begin{figure}[!t]
    \centering
    \begin{tikzpicture}
        \begin{groupplot}[
            cycle list name=color list,
            no markers,
            legend columns=2,
            height=3.5cm,
            width = 0.6\linewidth,
            xlabel={\small Iter. (in thous.)},
            ylabel={\small Samp. Prob.},
            x label style={at={(0.5,0.15)}},
            y label style={at={(0.15,0.5)}},
            legend columns=4,
            legend style={at={(-0.15,-0.35)},anchor=north,font=\small},
            group style={
                group size=2 by 1,
                xlabels at=edge top,
                x descriptions at=edge bottom,
                y descriptions at=edge left,
                horizontal sep=0pt
            },
            tick label style={font=\tiny}
        ]
        \nextgroupplot[
            title={\small \dds},
        ]
        \addplot table [mark=none, x=idx, y=WMT, col sep=comma] {multidds-multidomain-probs.csv};
        \addplot table [mark=none, x=idx, y=Tanzil, col sep=comma] {multidds-multidomain-probs.csv};
        \addplot table [mark=none, x=idx, y=EMEA, col sep=comma] {multidds-multidomain-probs.csv};
        \addplot table [mark=none, x=idx, y=KDE, col sep=comma] {multidds-multidomain-probs.csv};
        
        \nextgroupplot[
            title={\small \ours},
        ]
        \addplot table [mark=none, x=idx, y=WMT, col sep=comma] {multiuat-multidomain-probs.csv};
        \addplot table [mark=none, x=idx, y=Tanzil, col sep=comma] {multiuat-multidomain-probs.csv};
        \addplot table [mark=none, x=idx, y=EMEA, col sep=comma] {multiuat-multidomain-probs.csv};
        \addplot table [mark=none, x=idx, y=KDE, col sep=comma] {multiuat-multidomain-probs.csv};
        \legend{\texttt{WMT}, \texttt{Tanzil}, \texttt{EMEA}, \texttt{KDE}}
        \end{groupplot}
        
    \end{tikzpicture}
    \caption{Iteration (Iter.)-sampling probability of \dds and \ours over multi-domain NMT training sets.}
    \label{fig:multidomaincurve}

\end{figure}

\subsection{Learned Distribution for Language Pairs/Domains}

We visualize the change of sampling distribution, w.r.t. the training iterations, of the multilingual O2M-diverse (\autoref{fig:multilingualcurve}) and multi-domain (\autoref{fig:multidomaincurve}) setting. In both figures, \dds and \ours gradually increase the usage of LRLs/LRDs and decrease the usage of HRLs/HRDs.
In the multilingual NMT, we observe that the learned distributions by both \dds and \ours converge from proportional sampling to uniform sampling with a mild trend to divergence in the one given by \dds. 
In the multi-domain NMT, \ours illustrates the consistent adjustment as the trend illustrated in multilingual O2M-diverse setting, but the learned distribution given by \dds is overwhelmed by \texttt{Tanzil}.

The model uncertainty focuses on how well the dataset is described by the model $\vtheta$, instead of the interference among datasets, so that \ours is free from the assumption on the cross-corpus transference and not affected by \texttt{Tanzil}.

\subsection{Why Cosine Similarity Fails?\footnote{In our preliminary study, we investigate various similarity-based rewards, such as dot product, but we do not observe significant difference.}}
\label{sec:whyfail}

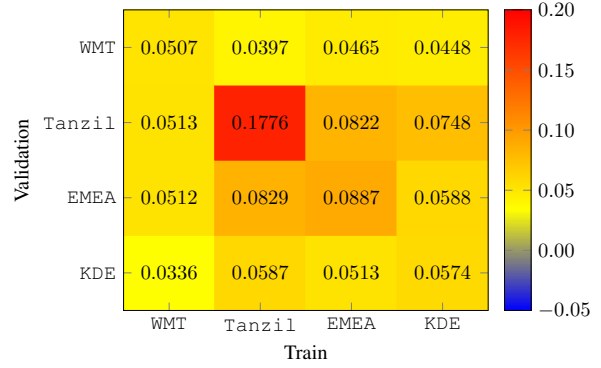
\begin{figure}[!t]
    \centering
    \begin{tikzpicture}[scale=0.7]
        \begin{axis}[
            nodes near coords,
            every node near coord/.append style={
                /pgf/number format/fixed zerofill,
                /pgf/number format/precision=4
            },
            nodes near coords align={center},
            view={0}{90},   
            xlabel=Train,
            ylabel=Validation,
            y label style={at={(-0.1,0.5)}},
            xticklabels={0, \texttt{WMT}, \texttt{Tanzil}, \texttt{EMEA}, \texttt{KDE}},
            yticklabels={0, \texttt{KDE}, \texttt{EMEA}, \texttt{Tanzil}, \texttt{WMT}},
            colorbar,
            colorbar style={
                yticklabel style={
                    /pgf/number format/.cd,
                    fixed,
                    precision=2,
                    fixed zerofill
                },
            },
            enlargelimits=false,
            axis on top,
            point meta min=-0.05,
            point meta max=0.2,
        ]
            \addplot [matrix plot*,point meta=explicit] file [meta=index 2] {cosine.dat};

        \end{axis}
    
    \end{tikzpicture}
    \caption{Averaged cosine similarity matrix between train and validation mini-batch during the optimization of \dds.}
    \label{fig:cosmat}
\end{figure}

\begin{table*}[t]
    \small
    \setlength{\tabcolsep}{8pt}
    \centering
    \begin{tabular}{lccccccccc}
        & & \multicolumn{5}{c}{ID}                                                    & \multicolumn{3}{c}{OOD}               \\ \cmidrule(lr){3-7} \cmidrule(l){8-10}
        & {$\tau$} & $\avgbleu$        & \texttt{WMT}            & \texttt{Tanzil}         & \texttt{EMEA}           & \texttt{KDE}           & $\avgbleu$        & \texttt{QED}            & \texttt{TED}            \\ 
        \toprule
        \dds &
        $1$      & 36.42          & \underline{21.76}    & 37.41          & 51.92          & 34.60           & \underline{22.74}    & \underline{20.30}     & \underline{25.18}    \\ 
        & $5$      & 36.16          & 19.50          & 37.55          & 52.04          & 35.56          & 21.60          & 19.59          & 23.61          \\ 
        & $\infty$    & 36.03          & 18.90          & 37.45          & 51.91          & \underline{35.85}    & 21.47          & 19.37          & 23.57          \\ \midrule
        \ours
        & $1$      & \textbf{37.44} & \textbf{22.14} & 38.38\rlap{$\dagger$}          & \textbf{52.81}\rlap{$\dagger$} & \textbf{36.44}\rlap{$\dagger$}  & \textbf{23.80} & \textbf{21.34}\rlap{$\dagger$} & \textbf{26.26}\rlap{$\dagger$} \\ 
        & $5$      & \underline{37.04}    & 20.34          & \underline{39.40}\rlap{$\dagger$}    & \underline{52.68}    & 35.73          & 22.46          & 19.97          & 24.95          \\ 
        & $\infty$    & 36.93          & 20.09          & \textbf{39.84}\rlap{$\dagger$} & 52.03          & 35.74          & 22.14          & 19.79          & 24.48          \\ \bottomrule
        \end{tabular}
    \caption{Effects of sampling priors ($\tau$) for \dds and \ours in multi-domain NMT. 
    Best results are highlighted in \textbf{bold} and second best results are highlighted in \underline{underline}.
    $\dagger$ indicates the improvement for the corresponding result against best \dds result is statistically significant at $p<0.05$ given by \citet{koehn-2004-statistical}.}
    \label{tab:priors}
\end{table*}

\begin{table}[t]
    \small
        \centering
        \begin{tabular}{lcccc}
               & \texttt{WMT}            & \texttt{Tanzil}         & \texttt{EMEA}           & \texttt{KDE}            \\
        \toprule
        \texttt{WMT}    & \textbf{25.09} & 12.76          & 23.60           & 24.95          \\ 
        \texttt{Tanzil} & \phantom{0}0.22           & \textbf{40.66} & \phantom{0}0.07           & \phantom{0}0.14           \\ 
        \texttt{EMEA}   & \phantom{0}4.78           & \phantom{0}1.39           & \textbf{54.25} & \phantom{0}7.01           \\ 
        \texttt{KDE}    & \phantom{0}4.19           & \phantom{0}1.74           & \phantom{0}9.54           & \textbf{30.71} \\ \bottomrule
        \end{tabular}
        \caption{The cross-domain evaluation for the single-domain NMT models. 
        BLEU scores in each row are produced by the corresponding single-domain NMT model. 
        Best results are highlighted in \textbf{bold}.}
        \label{tab:transfer}
\end{table}

A natural question is raised after seeing \autoref{fig:multidomaincurve}: \textit{why does Tanzil overwhelm the sampling distribution by \dds in multi-domain NMT?}

As in \autoref{eq:cosinereward}, \dds computes pairwise cosine similarities for all the language pairs/domains using sampled mini-batches between $\dtrain$ and $\dvalid$ to update the sampling probability.
We average all the cosine similarity matrices during the training and visualize the averaged matrix in \autoref{fig:cosmat}.
As visualized, \texttt{Tanzil} is a highly self-correlated domain whose cosine similarity is about at least two times larger than the other values in the matrix.
This leads to a very high reward on \texttt{Tanzil}, and the sampling probability of \texttt{Tanzil} in \dds keeps increasing to more than 40\% in \autoref{fig:multidomaincurve}. 

However, is \texttt{Tanzil} highly beneficial to the overall performance?
To probe the cross-domain generalization, we train four single-domain NMT models on each in-domain corpus and evaluate these models on all the in-domain test sets, and the results are presented in \autoref{tab:transfer}.
We can observe that the knowledge learned from \texttt{WMT} can be generalized to other domains, but the knowledge learned from \texttt{Tanzil} is almost not beneficial to other domains. 

Therefore, \dds with the data-dependent cosine similarity reward is vulnerable to the change of datasets and can be possibly overwhelmed by a special dataset like \texttt{Tanzil}, since the cross-corpus transfer is intractable.

\subsection{Effects of Sampling Priors}

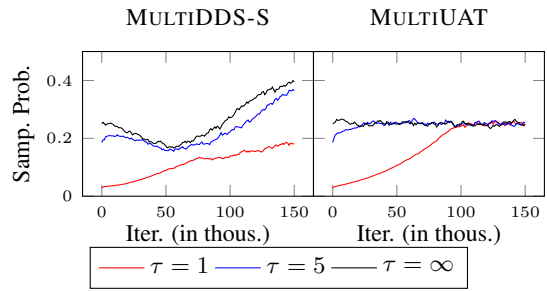
\begin{figure}[t]
    \centering
    \begin{tikzpicture}
        \begin{groupplot}[
            cycle list name=color list,
            no markers,
            legend columns=2,
            height=3.5cm,
            width = 0.6\linewidth,
            xlabel={\small Iter. (in thous.)},
            ylabel={\small Samp. Prob.},
            x label style={at={(0.5,0.15)}},
            y label style={at={(0.15,0.5)}},
            legend columns=4,
            legend style={at={(-0.15,-0.35)},anchor=north,font=\small},
            ymin=0, ymax=0.5,
            group style={
                group size=2 by 1,
                xlabels at=edge top,
                x descriptions at=edge bottom,
                y descriptions at=edge left,
                horizontal sep=0pt
            },
            tick label style={font=\tiny}
        ]
        \nextgroupplot[
            title={\small \dds},
        ]
        \addplot table [mark=none, x=idx, y=KDE4, col sep=comma] {multidds-multidomain-1-probs.csv};
        \addplot table [mark=none, x=idx, y=KDE4, col sep=comma] {multidds-multidomain-5-probs.csv};
        \addplot table [mark=none, x=idx, y=KDE4, col sep=comma] {multidds-multidomain-1e20-probs.csv};
        \nextgroupplot[
            title={\small \ours},
        ]
        \addplot table [mark=none, x=idx, y=KDE4, col sep=comma] {multiuat-multidomain-1-probs.csv};
        \addplot table [mark=none, x=idx, y=KDE4, col sep=comma] {multiuat-multidomain-5-probs.csv};
        \addplot table [mark=none, x=idx, y=KDE4, col sep=comma] {multiuat-multidomain-1e20-probs.csv};
        \legend{$\tau=1$,$\tau=5$,$\tau=\infty$}
        \end{groupplot}
        
    \end{tikzpicture}
    \caption{Iteration (Iter.)-sampling probability over \texttt{KDE} of \dds and \ours with different temperature priors.}
    \label{fig:diffprior}

\end{figure}

Both \dds and \ours initialize the sampling probability distribution to proportional distribution (\autoref{line:line1} in \autoref{alg:multiuat}). 
We investigate how the prior sampling distribution affects the performance and present the results in \autoref{tab:priors}. 
We can observe that the prior sampling distribution can affect the overall performance. 
For both \dds and \ours, the overall results on both in-domain and out-of-domain evaluation are negatively correlated with the prior $\tau$.

We also visualize the change of sampling probability of \texttt{KDE} given by \dds and \ours with different prior sampling distributions in \autoref{fig:diffprior}. The learned sampling distribution by \ours always converges to uniform distribution, regardless of the change of prior sampling distribution. However, the change of priors significantly affects the learned sampling distribution of \dds.

%% file: 7_related_work.tex
\section{Related Work}


\paragraph{Multi-corpus NLP} Multilingual training has been particularly prominent in recent advances driven by the demand of training a unified model for all the languages \citep{dong-etal-2015-multi, plank-etal-2016-multilingual, johnson-etal-2017-googles, DBLP:journals/corr/abs-1907-05019}. 
\citet{freitag-firat-2020-complete} extend current English-centric training to a many-to-many setup without sacrificing the performance on English-centric language pairs.
\citet{wang2021gradient} improve the multilingual training by adjusting gradient directions based on gradient similarity.
Existing works on multi-domain training commonly attempt to leverage architectural domain-specific components or auxiliary loss \citep{DBLP:journals/corr/abs-1708-08712, DBLP:journals/corr/abs-1805-02282, zeng-etal-2018-multi, li-etal-2018-whats, DBLP:conf/emnlp/DengYYDL20, jiang-etal-2020-multi}. 
These approaches commonly do not explore much on the training proportion across domains and are limited to in-domain prediction and less generalizable to unseen domains.
\citet{zaremoodi-haffari-2019-adaptively} dynamically balance the importance of tasks in multi-task NMT to improve the low-resource NMT performance.
\citet{vu-etal-2021-generalised} leverage a pre-trained language model to select useful monolingual data from either source language or target language to perform unsupervised domain adaptation for NMT models.
Our work is directly related to \citet{DBLP:conf/icml/WangPMACN20} and \citet{wang-etal-2020-balancing} that leverage cosine similarity of gradients as a reward to dynamically adjust the data usage in the multilingual training.

\paragraph{Model uncertainty} Estimating the sequence and word-level uncertainty via Monta Carlo Dropout \citep{DBLP:conf/icml/GalG16} has been investigated for NMT \citep{DBLP:journals/corr/abs-2006-08344, wang-etal-2019-improving-back, fomicheva-etal-2020-unsupervised,malinin2021uncertainty}. 
\citet{wang-etal-2019-improving-back} exploit model uncertainty on back-translation to reduce the noise in the back-translated corpus.
\citet{DBLP:journals/corr/abs-2006-08344} and \citet{malinin2021uncertainty} investigate to leverage model uncertainty to detect out-of-distribution translations.
\citet{fomicheva-etal-2020-unsupervised} summarize several measures to estimate quality of translated sentences, including the model uncertainty.
Our work exploits the uncertainty measures as suggested by \citet{wang-etal-2019-improving-back} and \citet{malinin2021uncertainty}.

%% file: 8_conclusion.tex
\section{Conclusion}

In this work, we propose \ours, a general model-agnostic framework that learns to automatically balance the data usage to achieve better overall performance on multiple corpora based on model uncertainty.
We run extensive experiments on both multilingual and multi-domain NMT, and empirically demonstrate the effectiveness of our approach.
Our approach substantially outperforms other baseline approaches.
We empirically point out the vulnerability of a comparable approach \dds \citep{wang-etal-2020-balancing}.

We focus on the problem that dynamically balances text corpora collected from heterogeneous sources in this paper.
However, the heterogeneity of text corpora is far beyond the languages and domains which are discussed in this work.
For example, the quality of datasets is not covered.
We leave the study on the quality of datasets to the future work.

\section*{Acknowledgments}
We would like to thank the anonymous reviewers for providing constructive and helpful comments on this work. This work is partly supported by the ARC Future Fellowship FT190100039.

%% file: appendix_1_results.tex
\section{Complete Results}\label{appsec:comp}

We present the complete results of our own implementation in \autoref{tab:multilingual-m2o-related}, \autoref{tab:multilingual-m2o-diverse}, \autoref{tab:multilingual-o2m-related}, \autoref{tab:multilingual-o2m-diverse} and \autoref{tab:multidomain-inout}. The multilingual results for heuristic static strategies and \dds are obtained with the hyperparameters provided by \citet{wang-etal-2020-balancing}.

\begin{table*}[]
\small
\centering
\begin{tabular}{@{}lccccccccc@{}}
\toprule
                               & $\avgbleu$ & \texttt{aze}   & \texttt{bel}   & \texttt{glg}   & \texttt{slk}   & \texttt{tur}   & \texttt{rus}   & \texttt{por}   & \texttt{ces}   \\ \midrule
\textsc{Prop.}\phantom{0}($\tau=1$)      & 24.78    & 10.76 & 16.75 & 28.24 & 28.61 & 22.89 & 22.93 & 41.53 & 26.54 \\ 
\textsc{Temp.} ($\tau=5$)      & 23.57    & \phantom{0}9.49  & 15.58 & 26.81 & 27.86 & 21.09 & 21.77 & 40.10 & 25.88 \\ 
\textsc{Uni.}\phantom{00}($\tau=\infty$)  & 22.41    & \phantom{0}8.03  & 14.26 & 24.82 & 27.38 & 19.99 & 20.93 & 38.78 & 25.08 \\ 
\dds                          & 24.68    & 11.35 & 17.70 & 28.36 & 29.08 & 21.62 & 22.15 & 40.68 & 26.50 \\ \midrule
\ours & & & & & & & & & \\
\phantom{000}+ \textsc{PreTP}   & 26.30    & 12.67\rlap{$\dagger$}  & 19.28\rlap{$\dagger$}  & 29.31\rlap{$\dagger$}  & 30.57\rlap{$\dagger$}  & 24.10\rlap{$\dagger$}  & 23.75\rlap{$\dagger$}  & 42.85\rlap{$\dagger$}  & 27.89\rlap{$\dagger$}  \\
\phantom{000}+ \textsc{ExpTP}   & 26.36    & 12.61\rlap{$\dagger$}  & 18.81\rlap{$\dagger$}  & 28.97  & \textbf{30.98}\rlap{$\dagger$}  & \textbf{24.56}\rlap{$\dagger$}  & \textbf{23.92}\rlap{$\dagger$}  & 42.83\rlap{$\dagger$}  & \textbf{28.22}\rlap{$\dagger$}  \\ 
\phantom{000}+ \textsc{VarTP}   & \textbf{26.39}    & 12.84\rlap{$\dagger$}  & \textbf{19.79}\rlap{$\dagger$}  & 29.43\rlap{$\dagger$}  & 30.63\rlap{$\dagger$}  & 24.12\rlap{$\dagger$}  & 23.68\rlap{$\dagger$}  & 42.88\rlap{$\dagger$}  & 27.77\rlap{$\dagger$}  \\ 
\phantom{000}+ \textsc{ComEV}   & 26.26    & \textbf{13.06}\rlap{$\dagger$}  & 19.40\rlap{$\dagger$}  & 29.31\rlap{$\dagger$}  & 30.47\rlap{$\dagger$}  & 23.81\rlap{$\dagger$}  & 23.57\rlap{$\dagger$}  & 42.77\rlap{$\dagger$}  & 27.71\rlap{$\dagger$}  \\ 
\phantom{000}+ \textsc{EntSent} & 26.34    & 12.38\rlap{$\dagger$}  & 19.75\rlap{$\dagger$}  & \textbf{29.53}\rlap{$\dagger$}  & 30.40\rlap{$\dagger$}  & 24.16\rlap{$\dagger$}  & 23.86\rlap{$\dagger$}  & \textbf{42.91}\rlap{$\dagger$}  & 27.69\rlap{$\dagger$}  \\ 
\phantom{000}+ \textsc{EntEOS}  & 26.28    & 12.72\rlap{$\dagger$}  & 19.22\rlap{$\dagger$}  & 29.40\rlap{$\dagger$}  & 30.43\rlap{$\dagger$}  & 24.02\rlap{$\dagger$}  & 23.77\rlap{$\dagger$}  & 42.68\rlap{$\dagger$}  & 28.03\rlap{$\dagger$}  \\ \bottomrule
\end{tabular}
\caption{The complete results for M2O-related setting in multilingual NMT. 
Best results are highlighted in \textbf{bold}.
$\dagger$ indicates the improvement for the corresponding \ours result against \dds result is statistically significant at $p<0.05$ given by \citet{koehn-2004-statistical}.}
\label{tab:multilingual-m2o-related}
\end{table*}

\begin{table*}[]
\small
\centering
\begin{tabular}{@{}lccccccccc@{}}
\toprule
                          & $\avgbleu$ & \texttt{bos}   & \texttt{mar}   & \texttt{hin}   & \texttt{mkd}   & \texttt{ell}   & \texttt{bul}   & \texttt{fra}   & \texttt{kor}   \\ \midrule
\textsc{Prop.}\phantom{0}($\tau=1$)      & 26.28      & 21.87 & \phantom{0}9.92  & 21.51 & 31.12 & 35.21 & 36.02 & 37.75 & 16.83 \\ 
\textsc{Temp.} ($\tau=5$)      & 25.77      & 23.18 & \phantom{0}9.72  & 21.01 & 30.61 & 34.29 & 34.98 & 36.49 & 15.90 \\ 
\textsc{Uni.}\phantom{00}($\tau=\infty$)  & 24.97      & 21.48 & \phantom{0}9.27  & 20.25 & 30.08 & 33.70 & 34.11 & 35.54 & 15.30 \\ 
\dds                       & 26.67      & 24.64 & 10.45 & 22.09 & 32.31 & 35.13 & 35.50 & 37.06 & 16.16 \\ \midrule
\ours & & & & & & & & &  \\
\phantom{000}+ \textsc{PreTP}   & 27.82      & \textbf{26.05}\rlap{$\dagger$}  & 11.08 & \textbf{23.72}\rlap{$\dagger$}  & 32.39 & 35.92\rlap{$\dagger$}  & 37.30\rlap{$\dagger$}  & 38.71\rlap{$\dagger$}  & 17.37\rlap{$\dagger$}  \\ 
\phantom{000}+ \textsc{ExpTP}   & 27.74      & 24.96 & \textbf{11.28}  & 23.60\rlap{$\dagger$}  & 32.84 & 35.92\rlap{$\dagger$}  & 37.08\rlap{$\dagger$}  & 38.77\rlap{$\dagger$}  & 17.47\rlap{$\dagger$}  \\ 
\phantom{000}+ \textsc{VarTP}   & \textbf{27.83}      & 25.45\rlap{$\dagger$}  & 10.84 & 23.42\rlap{$\dagger$}  & 32.89 & \textbf{36.28}\rlap{$\dagger$}  & \textbf{37.38}\rlap{$\dagger$}  & \textbf{38.89}\rlap{$\dagger$}  & \textbf{17.47}\rlap{$\dagger$}  \\ 
\phantom{000}+ \textsc{ComEV}   & 27.75      & 25.25 & 11.04 & 23.54\rlap{$\dagger$}  & 33.00\rlap{$\dagger$}  & 35.97\rlap{$\dagger$}  & 37.21\rlap{$\dagger$}  & 38.53\rlap{$\dagger$}  & 17.42\rlap{$\dagger$}  \\ 
\phantom{000}+ \textsc{EntSent} & 27.71      & 24.95 & 11.25  & 23.30\rlap{$\dagger$}  & 32.84 & 36.02\rlap{$\dagger$}  & 37.22\rlap{$\dagger$}  & 38.70\rlap{$\dagger$}  & 17.42\rlap{$\dagger$}  \\ 
\phantom{000}+ \textsc{EntEOS}  & 27.82      & 25.90\rlap{$\dagger$}  & 11.18  & 23.06\rlap{$\dagger$}  & \textbf{33.36}\rlap{$\dagger$}  & 35.90\rlap{$\dagger$}  & 37.26\rlap{$\dagger$}  & 38.50\rlap{$\dagger$}  & 17.38\rlap{$\dagger$}  \\ \bottomrule
\end{tabular}
\caption{The complete results for M2O-diverse setting in multilingual NMT. 
Best results are highlighted in \textbf{bold}.
$\dagger$ indicates the improvement for the corresponding \ours result against \dds result is statistically significant at $p<0.05$ given by \citet{koehn-2004-statistical}.}
\label{tab:multilingual-m2o-diverse}
\end{table*}

\begin{table*}[]
\small
\centering
\begin{tabular}{@{}lccccccccc@{}}
\toprule
                               & $\avgbleu$ & \texttt{aze}   & \texttt{bel}   & \texttt{glg}   & \texttt{slk}   & \texttt{tur}   & \texttt{rus}   & \texttt{por}   & \texttt{ces}   \\ \midrule
\textsc{Prop.}\phantom{0}($\tau=1$)     & 15.32      & 4.07 & \phantom{0}4.29  & 16.26 & 17.17 & 12.37 & 16.70 & 35.25 & 16.41 \\ 
\textsc{Temp.} ($\tau=5$)      & 16.44      & 6.66 & 11.12 & 21.56 & 18.51 & 10.67 & 14.57 & 32.24 & 16.18 \\ 
\textsc{Uni.}\phantom{00}($\tau=\infty$) & 15.26      & 5.91 & 10.02 & 20.64 & 17.55 & 9.45  & 13.11 & 30.32 & 15.07 \\ 
\dds                           & 16.64      & 6.93 & 12.01 & 22.51 & 18.47 & 10.69 & 14.58 & 32.04 & 15.85 \\ \midrule
\ours & & & & & & & & & \\
\phantom{000}+ \textsc{PreTP}   & 18.36      & 6.37 & 11.23 & 22.60 & \textbf{21.56}\rlap{$\dagger$}  & 13.41\rlap{$\dagger$}  & 17.15\rlap{$\dagger$}  & 35.88\rlap{$\dagger$}  & 18.66\rlap{$\dagger$}  \\ 
\phantom{000}+ \textsc{ExpTP}   & \textbf{18.64}      & 6.95 & 12.25 & 23.50\rlap{$\dagger$}  & 21.51\rlap{$\dagger$}  & 13.37\rlap{$\dagger$}  & \textbf{17.34}\rlap{$\dagger$}  & 35.86\rlap{$\dagger$}  & 18.30\rlap{$\dagger$}  \\ 
\phantom{000}+ \textsc{VarTP}   & 18.58      & 6.80 & 12.08 & \textbf{23.69}\rlap{$\dagger$}  & 21.08\rlap{$\dagger$}  & \textbf{13.64}\rlap{$\dagger$}  & 17.31\rlap{$\dagger$}  & 35.95\rlap{$\dagger$}  & 18.06\rlap{$\dagger$}  \\ 
\phantom{000}+ \textsc{ComEV}   & 18.58      & \textbf{7.10} & 12.04 & 23.43\rlap{$\dagger$}  & 21.20\rlap{$\dagger$}  & 13.32\rlap{$\dagger$}  & 17.17\rlap{$\dagger$}  & 35.90\rlap{$\dagger$}  & 18.49\rlap{$\dagger$}  \\
\phantom{000}+ \textsc{EntSent} & 18.63      & 6.80 & \textbf{12.39} & 23.52\rlap{$\dagger$}  & 21.37\rlap{$\dagger$}  & 13.54\rlap{$\dagger$}  & 17.31\rlap{$\dagger$}  & 35.86\rlap{$\dagger$}  & 18.22\rlap{$\dagger$}  \\
\phantom{000}+ \textsc{EntEOS}  & 18.56      & 6.41 & 11.95 & 23.05  & 21.70\rlap{$\dagger$}  & 13.39\rlap{$\dagger$}  & 17.17\rlap{$\dagger$}  & \textbf{36.09}\rlap{$\dagger$}  & \textbf{18.68}\rlap{$\dagger$}  \\ \bottomrule
\end{tabular}
\caption{The complete results for O2M-related setting in multilingual NMT. 
Best results are highlighted in \textbf{bold}.
$\dagger$ indicates the improvement for the corresponding \ours result against \dds result is statistically significant at $p<0.05$ given by \citet{koehn-2004-statistical}.}
\label{tab:multilingual-o2m-related}
\end{table*}

\begin{table*}[]
\small
\centering
\begin{tabular}{@{}lccccccccc@{}}
\toprule
                                & $\avgbleu$ & \texttt{bos}   & \texttt{mar}   & \texttt{hin}   & \texttt{mkd}   & \texttt{ell}   & \texttt{bul}   & \texttt{fra}   & \texttt{kor}  \\ \midrule
\textsc{Prop.}\phantom{0}($\tau=1$)    & 16.79      & \phantom{0}6.48  & 3.58 & 10.67 & 15.09 & 26.73 & 29.05 & 33.32 & \textbf{9.39} \\ 
\textsc{Temp.} ($\tau=5$)     & 18.18      & 14.61 & \textbf{4.94} & 14.65 & 20.56 & 24.66 & 27.39 & 29.92 & 8.69 \\ 
\textsc{Uni.}\phantom{00}($\tau=\infty$)  & 17.25      & 14.62 & 4.88 & 13.99 & 20.31 & 23.61 & 25.05 & 27.48 & 8.03 \\ 
\dds                           & 18.04      & \textbf{15.43} & 4.85 & 14.36 & 20.80 & 24.63 & 26.77 & 29.93 & 7.57 \\ \midrule
\ours & & & & & & & & & \\
\phantom{000}+ \textsc{PreTP}   & 19.57      & 14.88 & 4.71 & 14.66 & 22.36\rlap{$\dagger$}  & \textbf{27.21}\rlap{$\dagger$}  & 30.04\rlap{$\dagger$}  & 33.89\rlap{$\dagger$}  & 8.83\rlap{$\dagger$}  \\ 
\phantom{000}+ \textsc{ExpTP}   & 19.59      & 15.11 & 4.86 & 14.64 & 22.06\rlap{$\dagger$}  & 26.87\rlap{$\dagger$}  & 30.22\rlap{$\dagger$}  & 34.16\rlap{$\dagger$}  & 8.81\rlap{$\dagger$}  \\ 
\phantom{000}+ \textsc{VarTP}   & 19.62      & 15.35 & 4.67 & 14.52 & 22.13\rlap{$\dagger$}  & 26.45\rlap{$\dagger$}  & 30.43\rlap{$\dagger$}  & 34.36\rlap{$\dagger$}  & 9.01\rlap{$\dagger$}  \\ 
\phantom{000}+ \textsc{ComEV}   & 19.67      & 14.68 & 4.72 & 14.73 & \textbf{23.12}\rlap{$\dagger$}  & 27.10\rlap{$\dagger$}  & 30.23\rlap{$\dagger$}  & 34.14\rlap{$\dagger$}  & 8.62\rlap{$\dagger$}  \\ 
\phantom{000}+ \textsc{EntSent} & 19.68      & 14.51 & 4.69 & \textbf{14.74} & 22.70\rlap{$\dagger$}  & 26.72\rlap{$\dagger$}  & 30.52\rlap{$\dagger$}  & 34.45\rlap{$\dagger$}  & 9.09\rlap{$\dagger$}  \\ 
\phantom{000}+ \textsc{EntEOS}  & \textbf{19.76}      & 14.59 & 4.83 & 14.63 & 23.08\rlap{$\dagger$}  & 27.05\rlap{$\dagger$}  & \textbf{30.64}\rlap{$\dagger$}  & \textbf{34.61}\rlap{$\dagger$}  & 8.68\rlap{$\dagger$}  \\ \bottomrule
\end{tabular}
\caption{The complete results for O2M-diverse setting in multilingual NMT. 
Best results are highlighted in \textbf{bold}.
$\dagger$ indicates the improvement for the corresponding \ours result against \dds result is statistically significant at $p<0.05$ given by \citet{koehn-2004-statistical}.}
\label{tab:multilingual-o2m-diverse}
\end{table*}

\begin{table*}[]
\small
\centering
\begin{tabular}{@{}lcccccccc@{}}
\toprule
                                & \multicolumn{5}{c}{ID}                            & \multicolumn{3}{c}{OOD}   \\ \cmidrule(lr){2-6} \cmidrule(l){7-9}
                                & $\avgbleu$ & \texttt{WMT} & \texttt{Tanzil} & \texttt{EMEA}  & \texttt{KDE}   & $\avgbleu$ & \texttt{QED}   & \texttt{TED}   \\ \midrule
\textsc{Prop.}\phantom{0}($\tau=1$)   & 35.69      & \textbf{24.76}        & \textbf{40.23}  & 45.56 & 32.19 & \textbf{25.15}      & \textbf{22.71} & \textbf{27.59} \\ 
\textsc{Temp.} ($\tau=5$)      & 36.92      & 21.63        & 38.49  & 51.53 & 36.03 & 23.46      & 21.13 & 25.78 \\ 
\textsc{Uni.}\phantom{00}($\tau=\infty$)       & 36.85      & 20.17        & 38.89  & 52.17 & 36.16 & 22.22      & 19.85 & 24.58 \\ 
\dds                          & 36.42      & 21.76        & 37.41  & 51.92 & 34.60 & 22.74      & 20.30 & 25.18 \\ \midrule
\ours & & & & & & & &  \\
\phantom{000}+ \textsc{PreTP}   & \textbf{37.50}      & 22.21        & 39.61\rlap{$\dagger$}   & 52.66  & 35.53\rlap{$\dagger$}  & 23.78      & 21.20\rlap{$\dagger$}  & 26.35\rlap{$\dagger$}  \\ 
\phantom{000}+ \textsc{ExpTP}   & 37.29      & 22.22        & 39.37\rlap{$\dagger$}   & 51.62\rlap{$\dagger$} & 35.94\rlap{$\dagger$}  & 23.89      & 21.31\rlap{$\dagger$}  & 26.46\rlap{$\dagger$}  \\ 
\phantom{000}+ \textsc{VarTP}   & 35.48      & 23.08\rlap{$\dagger$}         & 39.11\rlap{$\dagger$}   & 46.91 & 32.80 & 23.75      & 21.21\rlap{$\dagger$}  & 26.29\rlap{$\dagger$}  \\ 
\phantom{000}+ \textsc{CombTP}  & 37.25      & 21.73        & 38.51\rlap{$\dagger$}   & \textbf{52.83}\rlap{$\dagger$}  & 35.91\rlap{$\dagger$}  & 23.36      & 20.98\rlap{$\dagger$}  & 25.74 \\ 
\phantom{000}+ \textsc{EntSent} & 37.27      & 22.64\rlap{$\dagger$}         & 38.64\rlap{$\dagger$}   & 51.65 & 36.14\rlap{$\dagger$}  & 23.86      & 21.29\rlap{$\dagger$}  & 26.42\rlap{$\dagger$}  \\ 
\phantom{000}+ \textsc{EntEOS}  & 37.44      & 22.14        & 38.38\rlap{$\dagger$}   & 52.81\rlap{$\dagger$}  & \textbf{36.44}\rlap{$\dagger$}  & 23.80      & 21.34\rlap{$\dagger$}  & 26.26\rlap{$\dagger$}  \\ \bottomrule
\end{tabular}
\caption{The complete results for multi-domain NMT. 
Best results are highlighted in \textbf{bold}.
$\dagger$ indicates the improvement for the corresponding \ours result against \dds result is statistically significant at $p<0.05$ given by \citet{koehn-2004-statistical}.}
\label{tab:multidomain-inout}
\end{table*}

%% file: appendix_2_hyperparams.tex
\section{Hyperparameters for Optimization}
\paragraph{Multilingual NMT} For \ours, the NMT model is optimized with Adam \cite{DBLP:journals/corr/KingmaB14} with $\beta_{1}=0.9$ and $\beta_{2}=0.98$. 
The model is optimized for $40$ epochs with the learning rate $\alpha=5 \times 10^{-4}$ and the batch size of $9600$ tokens. 
The learning rate increases linearly in the first $4K$ steps to the peak and then declines proportionally to the inverse square root of the number of steps. 
$\vscorer$ is updated for every $2K$ steps with the learning rate $1 \times 10^{-4}$.

\paragraph{Multi-domain NMT} The NMT model is optimized with Adam \cite{DBLP:journals/corr/KingmaB14} with $\beta_{1}=0.9$ and $\beta_{2}=0.98$. 
The model is optimized for $20$ epochs with the learning rate $\alpha=7 \times 10^{-4}$ and the batch size of $32K$ tokens. 
The learning rate increases linearly in the first $4K$ steps to the peak and then declines proportionally to the inverse square root of the number of steps. 
$\vscorer$ for both \dds and \ours is updated for every $1K$ steps with the learning rate $1 \times 10^{-4}$. 
All the hyperparameters are identical among all the approaches in multi-domain setup.

%% file: emnlp2021.bbl
\begin{thebibliography}{49}
\expandafter\ifx\csname natexlab\endcsname\relax\def\natexlab#1{#1}\fi

\bibitem[{Abdelali et~al.(2014)Abdelali, Guzman, Sajjad, and
  Vogel}]{abdelali-etal-2014-amara}
Ahmed Abdelali, Francisco Guzman, Hassan Sajjad, and Stephan Vogel. 2014.
\newblock \href
  {http://www.lrec-conf.org/proceedings/lrec2014/pdf/877_Paper.pdf} {The
  {AMARA} corpus: Building parallel language resources for the educational
  domain}.
\newblock In \emph{Proceedings of the Ninth International Conference on
  Language Resources and Evaluation ({LREC}'14)}, pages 1856--1862, Reykjavik,
  Iceland. European Language Resources Association (ELRA).

\bibitem[{Aharoni et~al.(2019)Aharoni, Johnson, and
  Firat}]{aharoni-etal-2019-massively}
Roee Aharoni, Melvin Johnson, and Orhan Firat. 2019.
\newblock \href {https://doi.org/10.18653/v1/N19-1388} {Massively multilingual
  neural machine translation}.
\newblock In \emph{Proceedings of the 2019 Conference of the North {A}merican
  Chapter of the Association for Computational Linguistics: Human Language
  Technologies, Volume 1 (Long and Short Papers)}, pages 3874--3884,
  Minneapolis, Minnesota. Association for Computational Linguistics.

\bibitem[{Arivazhagan et~al.(2019)Arivazhagan, Bapna, Firat, Lepikhin, Johnson,
  Krikun, Chen, Cao, Foster, Cherry, Macherey, Chen, and
  Wu}]{DBLP:journals/corr/abs-1907-05019}
Naveen Arivazhagan, Ankur Bapna, Orhan Firat, Dmitry Lepikhin, Melvin Johnson,
  Maxim Krikun, Mia~Xu Chen, Yuan Cao, George~F. Foster, Colin Cherry, Wolfgang
  Macherey, Zhifeng Chen, and Yonghui Wu. 2019.
\newblock \href {http://arxiv.org/abs/1907.05019} {Massively multilingual
  neural machine translation in the wild: Findings and challenges}.
\newblock \emph{CoRR}, abs/1907.05019.

\bibitem[{Arthur et~al.(2021)Arthur, Ryu, and
  Haffari}]{arthur-etal-2021-multilingual}
Philip Arthur, Dongwon Ryu, and Gholamreza Haffari. 2021.
\newblock \href {https://doi.org/10.18653/v1/2021.findings-acl.420}
  {Multilingual simultaneous neural machine translation}.
\newblock In \emph{Findings of the Association for Computational Linguistics:
  ACL-IJCNLP 2021}, pages 4758--4766, Online. Association for Computational
  Linguistics.

\bibitem[{Blundell et~al.(2015)Blundell, Cornebise, Kavukcuoglu, and
  Wierstra}]{pmlr-v37-blundell15}
Charles Blundell, Julien Cornebise, Koray Kavukcuoglu, and Daan Wierstra. 2015.
\newblock \href {http://proceedings.mlr.press/v37/blundell15.html} {Weight
  uncertainty in neural network}.
\newblock In \emph{Proceedings of the 32nd International Conference on Machine
  Learning}, volume~37 of \emph{Proceedings of Machine Learning Research},
  pages 1613--1622, Lille, France. PMLR.

\bibitem[{Bojar et~al.(2014)Bojar, Buck, Federmann, Haddow, Koehn, Leveling,
  Monz, Pecina, Post, Saint-Amand, Soricut, Specia, and
  Tamchyna}]{bojar-etal-2014-findings}
Ond{\v{r}}ej Bojar, Christian Buck, Christian Federmann, Barry Haddow, Philipp
  Koehn, Johannes Leveling, Christof Monz, Pavel Pecina, Matt Post, Herve
  Saint-Amand, Radu Soricut, Lucia Specia, and Ale{\v{s}} Tamchyna. 2014.
\newblock \href {https://doi.org/10.3115/v1/W14-3302} {Findings of the 2014
  workshop on statistical machine translation}.
\newblock In \emph{Proceedings of the Ninth Workshop on Statistical Machine
  Translation}, pages 12--58, Baltimore, Maryland, USA. Association for
  Computational Linguistics.

\bibitem[{Buntine and Weigend(1991)}]{DBLP:journals/compsys/BuntineW91}
Wray~L. Buntine and Andreas~S. Weigend. 1991.
\newblock \href {http://www.complex-systems.com/abstracts/v05\_i06\_a04.html}
  {Bayesian back-propagation}.
\newblock \emph{Complex Syst.}, 5(6).

\bibitem[{Colson et~al.(2007)Colson, Marcotte, and
  Savard}]{DBLP:journals/anor/ColsonMS07}
Beno{\^{\i}}t Colson, Patrice Marcotte, and Gilles Savard. 2007.
\newblock \href {https://doi.org/10.1007/s10479-007-0176-2} {An overview of
  bilevel optimization}.
\newblock \emph{Ann. Oper. Res.}, 153(1):235--256.

\bibitem[{Conneau et~al.(2020)Conneau, Khandelwal, Goyal, Chaudhary, Wenzek,
  Guzm{\'a}n, Grave, Ott, Zettlemoyer, and
  Stoyanov}]{conneau-etal-2020-unsupervised}
Alexis Conneau, Kartikay Khandelwal, Naman Goyal, Vishrav Chaudhary, Guillaume
  Wenzek, Francisco Guzm{\'a}n, Edouard Grave, Myle Ott, Luke Zettlemoyer, and
  Veselin Stoyanov. 2020.
\newblock \href {https://doi.org/10.18653/v1/2020.acl-main.747} {Unsupervised
  cross-lingual representation learning at scale}.
\newblock In \emph{Proceedings of the 58th Annual Meeting of the Association
  for Computational Linguistics}, pages 8440--8451, Online. Association for
  Computational Linguistics.

\bibitem[{Daum{\'e}~III(2007)}]{daume-iii-2007-frustratingly}
Hal Daum{\'e}~III. 2007.
\newblock \href {https://www.aclweb.org/anthology/P07-1033} {Frustratingly easy
  domain adaptation}.
\newblock In \emph{Proceedings of the 45th Annual Meeting of the Association of
  Computational Linguistics}, pages 256--263, Prague, Czech Republic.
  Association for Computational Linguistics.

\bibitem[{Deng et~al.(2020)Deng, Yu, Yu, Duan, and
  Luo}]{DBLP:conf/emnlp/DengYYDL20}
Yongchao Deng, Hongfei Yu, Heng Yu, Xiangyu Duan, and Weihua Luo. 2020.
\newblock \href {https://doi.org/10.18653/v1/2020.findings-emnlp.377}
  {Factorized transformer for multi-domain neural machine translation}.
\newblock In \emph{Proceedings of the 2020 Conference on Empirical Methods in
  Natural Language Processing: Findings, {EMNLP} 2020, Online Event, 16-20
  November 2020}, pages 4221--4230. Association for Computational Linguistics.

\bibitem[{Dong et~al.(2015)Dong, Wu, He, Yu, and Wang}]{dong-etal-2015-multi}
Daxiang Dong, Hua Wu, Wei He, Dianhai Yu, and Haifeng Wang. 2015.
\newblock \href {https://doi.org/10.3115/v1/P15-1166} {Multi-task learning for
  multiple language translation}.
\newblock In \emph{Proceedings of the 53rd Annual Meeting of the Association
  for Computational Linguistics and the 7th International Joint Conference on
  Natural Language Processing (Volume 1: Long Papers)}, pages 1723--1732,
  Beijing, China. Association for Computational Linguistics.

\bibitem[{Dong et~al.(2018)Dong, Quirk, and Lapata}]{dong-etal-2018-confidence}
Li~Dong, Chris Quirk, and Mirella Lapata. 2018.
\newblock \href {https://doi.org/10.18653/v1/P18-1069} {Confidence modeling for
  neural semantic parsing}.
\newblock In \emph{Proceedings of the 56th Annual Meeting of the Association
  for Computational Linguistics (Volume 1: Long Papers)}, pages 743--753,
  Melbourne, Australia. Association for Computational Linguistics.

\bibitem[{Fang et~al.(2017)Fang, Li, and Cohn}]{DBLP:conf/emnlp/FangLC17}
Meng Fang, Yuan Li, and Trevor Cohn. 2017.
\newblock Learning how to active learn: {A} deep reinforcement learning
  approach.
\newblock In \emph{Proceedings of the 2017 Conference on Empirical Methods in
  Natural Language Processing, {EMNLP} 2017, Copenhagen, Denmark, September
  9-11, 2017}, pages 595--605. Association for Computational Linguistics.

\bibitem[{Fomicheva et~al.(2020)Fomicheva, Sun, Yankovskaya, Blain, Guzm{\'a}n,
  Fishel, Aletras, Chaudhary, and Specia}]{fomicheva-etal-2020-unsupervised}
Marina Fomicheva, Shuo Sun, Lisa Yankovskaya, Fr{\'e}d{\'e}ric Blain, Francisco
  Guzm{\'a}n, Mark Fishel, Nikolaos Aletras, Vishrav Chaudhary, and Lucia
  Specia. 2020.
\newblock \href {https://doi.org/10.1162/tacl_a_00330} {Unsupervised quality
  estimation for neural machine translation}.
\newblock \emph{Transactions of the Association for Computational Linguistics},
  8:539--555.

\bibitem[{Freitag and Firat(2020)}]{freitag-firat-2020-complete}
Markus Freitag and Orhan Firat. 2020.
\newblock \href {https://www.aclweb.org/anthology/2020.wmt-1.66} {Complete
  multilingual neural machine translation}.
\newblock In \emph{Proceedings of the Fifth Conference on Machine Translation},
  pages 550--560, Online. Association for Computational Linguistics.

\bibitem[{Gal and Ghahramani(2016)}]{DBLP:conf/icml/GalG16}
Yarin Gal and Zoubin Ghahramani. 2016.
\newblock \href {http://proceedings.mlr.press/v48/gal16.html} {Dropout as a
  bayesian approximation: Representing model uncertainty in deep learning}.
\newblock In \emph{Proceedings of the 33nd International Conference on Machine
  Learning, {ICML} 2016, New York City, NY, USA, June 19-24, 2016}, volume~48
  of \emph{{JMLR} Workshop and Conference Proceedings}, pages 1050--1059.
  JMLR.org.

\bibitem[{Graves(2011)}]{DBLP:conf/nips/Graves11}
Alex Graves. 2011.
\newblock \href
  {https://proceedings.neurips.cc/paper/2011/hash/7eb3c8be3d411e8ebfab08eba5f49632-Abstract.html}
  {Practical variational inference for neural networks}.
\newblock In \emph{Advances in Neural Information Processing Systems 24: 25th
  Annual Conference on Neural Information Processing Systems 2011. Proceedings
  of a meeting held 12-14 December 2011, Granada, Spain}, pages 2348--2356.

\bibitem[{Jiang et~al.(2020)Jiang, Guo, Chen, Li, Xu, Lyu, and
  Zhu}]{jiang-etal-2020-multi}
Wenbin Jiang, Mengfei Guo, Yufeng Chen, Ying Li, Jinan Xu, Yajuan Lyu, and Yong
  Zhu. 2020.
\newblock \href {https://www.aclweb.org/anthology/2020.aacl-main.73}
  {Multi-view classification model for knowledge graph completion}.
\newblock In \emph{Proceedings of the 1st Conference of the Asia-Pacific
  Chapter of the Association for Computational Linguistics and the 10th
  International Joint Conference on Natural Language Processing}, pages
  726--734, Suzhou, China. Association for Computational Linguistics.

\bibitem[{Johnson et~al.(2017)Johnson, Schuster, Le, Krikun, Wu, Chen, Thorat,
  Vi{\'e}gas, Wattenberg, Corrado, Hughes, and
  Dean}]{johnson-etal-2017-googles}
Melvin Johnson, Mike Schuster, Quoc~V. Le, Maxim Krikun, Yonghui Wu, Zhifeng
  Chen, Nikhil Thorat, Fernanda Vi{\'e}gas, Martin Wattenberg, Greg Corrado,
  Macduff Hughes, and Jeffrey Dean. 2017.
\newblock \href {https://doi.org/10.1162/tacl_a_00065} {{G}oogle{'}s
  multilingual neural machine translation system: Enabling zero-shot
  translation}.
\newblock \emph{Transactions of the Association for Computational Linguistics},
  5:339--351.

\bibitem[{Kendall and Gal(2017)}]{DBLP:conf/nips/KendallG17}
Alex Kendall and Yarin Gal. 2017.
\newblock \href
  {https://proceedings.neurips.cc/paper/2017/hash/2650d6089a6d640c5e85b2b88265dc2b-Abstract.html}
  {What uncertainties do we need in bayesian deep learning for computer
  vision?}
\newblock In \emph{Advances in Neural Information Processing Systems 30: Annual
  Conference on Neural Information Processing Systems 2017, December 4-9, 2017,
  Long Beach, CA, {USA}}, pages 5574--5584.

\bibitem[{Kingma and Ba(2015)}]{DBLP:journals/corr/KingmaB14}
Diederik~P. Kingma and Jimmy Ba. 2015.
\newblock \href {http://arxiv.org/abs/1412.6980} {Adam: {A} method for
  stochastic optimization}.
\newblock In \emph{3rd International Conference on Learning Representations,
  {ICLR} 2015, San Diego, CA, USA, May 7-9, 2015, Conference Track
  Proceedings}.

\bibitem[{Koehn(2004)}]{koehn-2004-statistical}
Philipp Koehn. 2004.
\newblock \href {https://www.aclweb.org/anthology/W04-3250} {Statistical
  significance tests for machine translation evaluation}.
\newblock In \emph{Proceedings of the 2004 Conference on Empirical Methods in
  Natural Language Processing}, pages 388--395, Barcelona, Spain. Association
  for Computational Linguistics.

\bibitem[{Koehn et~al.(2007)Koehn, Hoang, Birch, Callison-Burch, Federico,
  Bertoldi, Cowan, Shen, Moran, Zens, Dyer, Bojar, Constantin, and
  Herbst}]{koehn-etal-2007-moses}
Philipp Koehn, Hieu Hoang, Alexandra Birch, Chris Callison-Burch, Marcello
  Federico, Nicola Bertoldi, Brooke Cowan, Wade Shen, Christine Moran, Richard
  Zens, Chris Dyer, Ond{\v{r}}ej Bojar, Alexandra Constantin, and Evan Herbst.
  2007.
\newblock \href {https://www.aclweb.org/anthology/P07-2045} {{M}oses: Open
  source toolkit for statistical machine translation}.
\newblock In \emph{Proceedings of the 45th Annual Meeting of the Association
  for Computational Linguistics Companion Volume Proceedings of the Demo and
  Poster Sessions}, pages 177--180, Prague, Czech Republic. Association for
  Computational Linguistics.

\bibitem[{Li et~al.(2018)Li, Baldwin, and Cohn}]{li-etal-2018-whats}
Yitong Li, Timothy Baldwin, and Trevor Cohn. 2018.
\newblock \href {https://doi.org/10.18653/v1/N18-2076} {What{'}s in a domain?
  learning domain-robust text representations using adversarial training}.
\newblock In \emph{Proceedings of the 2018 Conference of the North {A}merican
  Chapter of the Association for Computational Linguistics: Human Language
  Technologies, Volume 2 (Short Papers)}, pages 474--479, New Orleans,
  Louisiana. Association for Computational Linguistics.

\bibitem[{Li et~al.(2019)Li, Baldwin, and Cohn}]{li-etal-2019-semi}
Yitong Li, Timothy Baldwin, and Trevor Cohn. 2019.
\newblock \href {https://doi.org/10.18653/v1/P19-1186} {Semi-supervised
  stochastic multi-domain learning using variational inference}.
\newblock In \emph{Proceedings of the 57th Annual Meeting of the Association
  for Computational Linguistics}, pages 1923--1934, Florence, Italy.
  Association for Computational Linguistics.

\bibitem[{Malinin and Gales(2021)}]{malinin2021uncertainty}
Andrey Malinin and Mark Gales. 2021.
\newblock \href {https://openreview.net/forum?id=jN5y-zb5Q7m} {Uncertainty
  estimation in autoregressive structured prediction}.
\newblock In \emph{International Conference on Learning Representations}.

\bibitem[{Ott et~al.(2019)Ott, Edunov, Baevski, Fan, Gross, Ng, Grangier, and
  Auli}]{ott-etal-2019-fairseq}
Myle Ott, Sergey Edunov, Alexei Baevski, Angela Fan, Sam Gross, Nathan Ng,
  David Grangier, and Michael Auli. 2019.
\newblock \href {https://doi.org/10.18653/v1/N19-4009} {fairseq: A fast,
  extensible toolkit for sequence modeling}.
\newblock In \emph{Proceedings of the 2019 Conference of the North {A}merican
  Chapter of the Association for Computational Linguistics (Demonstrations)},
  pages 48--53, Minneapolis, Minnesota. Association for Computational
  Linguistics.

\bibitem[{Papineni et~al.(2002)Papineni, Roukos, Ward, and
  Zhu}]{papineni-etal-2002-bleu}
Kishore Papineni, Salim Roukos, Todd Ward, and Wei-Jing Zhu. 2002.
\newblock \href {https://doi.org/10.3115/1073083.1073135} {{B}leu: a method for
  automatic evaluation of machine translation}.
\newblock In \emph{Proceedings of the 40th Annual Meeting of the Association
  for Computational Linguistics}, pages 311--318, Philadelphia, Pennsylvania,
  USA. Association for Computational Linguistics.

\bibitem[{Plank et~al.(2016)Plank, S{\o}gaard, and
  Goldberg}]{plank-etal-2016-multilingual}
Barbara Plank, Anders S{\o}gaard, and Yoav Goldberg. 2016.
\newblock \href {https://doi.org/10.18653/v1/P16-2067} {Multilingual
  part-of-speech tagging with bidirectional long short-term memory models and
  auxiliary loss}.
\newblock In \emph{Proceedings of the 54th Annual Meeting of the Association
  for Computational Linguistics (Volume 2: Short Papers)}, pages 412--418,
  Berlin, Germany. Association for Computational Linguistics.

\bibitem[{Post(2018)}]{post-2018-call}
Matt Post. 2018.
\newblock \href {https://doi.org/10.18653/v1/W18-6319} {A call for clarity in
  reporting {BLEU} scores}.
\newblock In \emph{Proceedings of the Third Conference on Machine Translation:
  Research Papers}, pages 186--191, Brussels, Belgium. Association for
  Computational Linguistics.

\bibitem[{Sajjad et~al.(2017)Sajjad, Durrani, Dalvi, Belinkov, and
  Vogel}]{DBLP:journals/corr/abs-1708-08712}
Hassan Sajjad, Nadir Durrani, Fahim Dalvi, Yonatan Belinkov, and Stephan Vogel.
  2017.
\newblock \href {http://arxiv.org/abs/1708.08712} {Neural machine translation
  training in a multi-domain scenario}.
\newblock \emph{CoRR}, abs/1708.08712.

\bibitem[{Sennrich et~al.(2016)Sennrich, Haddow, and
  Birch}]{sennrich-etal-2016-neural}
Rico Sennrich, Barry Haddow, and Alexandra Birch. 2016.
\newblock \href {https://doi.org/10.18653/v1/P16-1162} {Neural machine
  translation of rare words with subword units}.
\newblock In \emph{Proceedings of the 54th Annual Meeting of the Association
  for Computational Linguistics (Volume 1: Long Papers)}, pages 1715--1725,
  Berlin, Germany. Association for Computational Linguistics.

\bibitem[{Tars and Fishel(2018)}]{DBLP:journals/corr/abs-1805-02282}
Sander Tars and Mark Fishel. 2018.
\newblock \href {http://arxiv.org/abs/1805.02282} {Multi-domain neural machine
  translation}.
\newblock \emph{CoRR}, abs/1805.02282.

\bibitem[{Tiedemann(2012)}]{tiedemann-2012-parallel}
J{\"o}rg Tiedemann. 2012.
\newblock \href
  {http://www.lrec-conf.org/proceedings/lrec2012/pdf/463_Paper.pdf} {Parallel
  data, tools and interfaces in {OPUS}}.
\newblock In \emph{Proceedings of the Eighth International Conference on
  Language Resources and Evaluation ({LREC}'12)}, pages 2214--2218, Istanbul,
  Turkey. European Language Resources Association (ELRA).

\bibitem[{Vaswani et~al.(2017)Vaswani, Shazeer, Parmar, Uszkoreit, Jones,
  Gomez, Kaiser, and Polosukhin}]{DBLP:conf/nips/VaswaniSPUJGKP17}
Ashish Vaswani, Noam Shazeer, Niki Parmar, Jakob Uszkoreit, Llion Jones,
  Aidan~N. Gomez, Lukasz Kaiser, and Illia Polosukhin. 2017.
\newblock \href
  {https://proceedings.neurips.cc/paper/2017/hash/3f5ee243547dee91fbd053c1c4a845aa-Abstract.html}
  {Attention is all you need}.
\newblock In \emph{Advances in Neural Information Processing Systems 30: Annual
  Conference on Neural Information Processing Systems 2017, December 4-9, 2017,
  Long Beach, CA, {USA}}, pages 5998--6008.

\bibitem[{von Stackelberg et~al.(2011)}]{von2011market}
Heinrich von Stackelberg et~al. 2011.
\newblock Market structure and equilibrium.
\newblock \emph{Springer Books}.

\bibitem[{Vu et~al.(2021)Vu, He, Phung, and Haffari}]{vu-etal-2021-generalised}
Thuy-Trang Vu, Xuanli He, Dinh Phung, and Gholamreza Haffari. 2021.
\newblock Generalised unsupervised domain adaptation of neural machine
  translation with cross-lingual data selection.
\newblock In \emph{Proceedings of the 2021 Conference on Empirical Methods in
  Natural Language Processing (EMNLP)}, Punta Cana, Dominican Republic.

\bibitem[{Wang et~al.(2019)Wang, Liu, Wang, Luan, and
  Sun}]{wang-etal-2019-improving-back}
Shuo Wang, Yang Liu, Chao Wang, Huanbo Luan, and Maosong Sun. 2019.
\newblock \href {https://doi.org/10.18653/v1/D19-1073} {Improving
  back-translation with uncertainty-based confidence estimation}.
\newblock In \emph{Proceedings of the 2019 Conference on Empirical Methods in
  Natural Language Processing and the 9th International Joint Conference on
  Natural Language Processing (EMNLP-IJCNLP)}, pages 791--802, Hong Kong,
  China. Association for Computational Linguistics.

\bibitem[{Wang and Neubig(2019)}]{wang-neubig-2019-target}
Xinyi Wang and Graham Neubig. 2019.
\newblock \href {https://doi.org/10.18653/v1/P19-1583} {Target conditioned
  sampling: Optimizing data selection for multilingual neural machine
  translation}.
\newblock In \emph{Proceedings of the 57th Annual Meeting of the Association
  for Computational Linguistics}, pages 5823--5828, Florence, Italy.
  Association for Computational Linguistics.

\bibitem[{Wang et~al.(2020{\natexlab{a}})Wang, Pham, Michel, Anastasopoulos,
  Carbonell, and Neubig}]{DBLP:conf/icml/WangPMACN20}
Xinyi Wang, Hieu Pham, Paul Michel, Antonios Anastasopoulos, Jaime~G.
  Carbonell, and Graham Neubig. 2020{\natexlab{a}}.
\newblock \href {http://proceedings.mlr.press/v119/wang20p.html} {Optimizing
  data usage via differentiable rewards}.
\newblock In \emph{Proceedings of the 37th International Conference on Machine
  Learning, {ICML} 2020, 13-18 July 2020, Virtual Event}, volume 119 of
  \emph{Proceedings of Machine Learning Research}, pages 9983--9995. {PMLR}.

\bibitem[{Wang et~al.(2020{\natexlab{b}})Wang, Tsvetkov, and
  Neubig}]{wang-etal-2020-balancing}
Xinyi Wang, Yulia Tsvetkov, and Graham Neubig. 2020{\natexlab{b}}.
\newblock \href {https://doi.org/10.18653/v1/2020.acl-main.754} {Balancing
  training for multilingual neural machine translation}.
\newblock In \emph{Proceedings of the 58th Annual Meeting of the Association
  for Computational Linguistics}, pages 8526--8537, Online. Association for
  Computational Linguistics.

\bibitem[{Wang et~al.(2021)Wang, Tsvetkov, Firat, and Cao}]{wang2021gradient}
Zirui Wang, Yulia Tsvetkov, Orhan Firat, and Yuan Cao. 2021.
\newblock \href {https://openreview.net/forum?id=F1vEjWK-lH_} {Gradient
  vaccine: Investigating and improving multi-task optimization in massively
  multilingual models}.
\newblock In \emph{International Conference on Learning Representations}.

\bibitem[{Williams(1992)}]{10.1007/BF00992696}
Ronald~J. Williams. 1992.
\newblock \href {https://doi.org/10.1007/BF00992696} {Simple statistical
  gradient-following algorithms for connectionist reinforcement learning}.
\newblock \emph{Mach. Learn.}, 8(3–4):229–256.

\bibitem[{Wu and Dredze(2019)}]{wu-dredze-2019-beto}
Shijie Wu and Mark Dredze. 2019.
\newblock \href {https://doi.org/10.18653/v1/D19-1077} {Beto, bentz, becas: The
  surprising cross-lingual effectiveness of {BERT}}.
\newblock In \emph{Proceedings of the 2019 Conference on Empirical Methods in
  Natural Language Processing and the 9th International Joint Conference on
  Natural Language Processing (EMNLP-IJCNLP)}, pages 833--844, Hong Kong,
  China. Association for Computational Linguistics.

\bibitem[{Xiao et~al.(2020)Xiao, Gomez, and
  Gal}]{DBLP:journals/corr/abs-2006-08344}
Tim~Z. Xiao, Aidan~N. Gomez, and Yarin Gal. 2020.
\newblock \href {http://arxiv.org/abs/2006.08344} {Wat zei je? detecting
  out-of-distribution translations with variational transformers}.
\newblock \emph{CoRR}, abs/2006.08344.

\bibitem[{Xiao and Wang(2019)}]{DBLP:conf/aaai/XiaoW19}
Yijun Xiao and William~Yang Wang. 2019.
\newblock \href {https://doi.org/10.1609/aaai.v33i01.33017322} {Quantifying
  uncertainties in natural language processing tasks}.
\newblock In \emph{The Thirty-Third {AAAI} Conference on Artificial
  Intelligence, {AAAI} 2019, The Thirty-First Innovative Applications of
  Artificial Intelligence Conference, {IAAI} 2019, The Ninth {AAAI} Symposium
  on Educational Advances in Artificial Intelligence, {EAAI} 2019, Honolulu,
  Hawaii, USA, January 27 - February 1, 2019}, pages 7322--7329. {AAAI} Press.

\bibitem[{Zaremoodi and Haffari(2019)}]{zaremoodi-haffari-2019-adaptively}
Poorya Zaremoodi and Gholamreza Haffari. 2019.
\newblock \href {https://doi.org/10.18653/v1/D19-5618} {Adaptively scheduled
  multitask learning: The case of low-resource neural machine translation}.
\newblock In \emph{Proceedings of the 3rd Workshop on Neural Generation and
  Translation}, pages 177--186, Hong Kong. Association for Computational
  Linguistics.

\bibitem[{Zeng et~al.(2018)Zeng, Su, Wen, Liu, Xie, Yin, and
  Zhao}]{zeng-etal-2018-multi}
Jiali Zeng, Jinsong Su, Huating Wen, Yang Liu, Jun Xie, Yongjing Yin, and
  Jianqiang Zhao. 2018.
\newblock \href {https://doi.org/10.18653/v1/D18-1041} {Multi-domain neural
  machine translation with word-level domain context discrimination}.
\newblock In \emph{Proceedings of the 2018 Conference on Empirical Methods in
  Natural Language Processing}, pages 447--457, Brussels, Belgium. Association
  for Computational Linguistics.

\end{thebibliography}
